\documentclass[lettersize,journal]{IEEEtran}
\usepackage[pdftex]{graphicx}
\usepackage{amsmath,amsfonts}
\usepackage{algorithmic}
\usepackage{algorithm}
\usepackage{array}
\usepackage[caption=false,font=normalsize,labelfont=sf,textfont=sf]{subfig}
\usepackage{textcomp}
\usepackage{stfloats}
\usepackage{url}
\usepackage{verbatim}
\usepackage{cite}

\usepackage{bm}
\usepackage{autobreak}

\begin{document}

\title{Study of Emotion Concept Formation by Integrating Vision, Physiology, and Word Information using Multilayered Multimodal Latent Dirichlet Allocation}
\author{Kazuki Tsurumaki, Chie Hieida and Kazuki Miyazawa

\thanks{K. Tsurumaki is with Graduate School of Engineering Science, Osaka University, Osaka, Japan}
\thanks{C. Hieida is with Graduate School of Science and Technology, Nara Institute of Science and Technology, Nara, Japan}
\thanks{K. Miyazawa is with Graduate School of Engineering Science, Osaka University, Osaka, Japan}
}

\markboth{Journal of \LaTeX\ Class Files,~Vol.~14, No.~8, August~2021}%
{Shell \MakeLowercase{\textit{et al.}}: A Sample Article Using IEEEtran.cls for IEEE Journals}

\IEEEpubid{0000--0000/00\$00.00~\copyright~2021 IEEE}

\maketitle

\begin{abstract}
How are emotions formed? Through extensive debate and the promulgation of diverse theories , the theory of constructed emotion has become prevalent in recent research on emotions. According to this theory, an emotion concept refers to a category  formed by interoceptive and exteroceptive information associated with a specific emotion. An emotion concept stores past experiences as knowledge and can predict unobserved information from acquired information. Therefore, in this study, we attempted to model the formation of emotion concepts using a constructionist approach from the perspective of the constructed emotion theory. Particularly, we constructed a model using multilayered multimodal latent Dirichlet allocation , which is a probabilistic generative model. We then trained the model for each subject using vision, physiology, and word information obtained from multiple people who experienced different visual emotion-evoking stimuli. To evaluate the model, we verified whether the formed categories matched human subjectivity and determined whether unobserved information could be predicted via categories. 
The verification results exceeded chance level, suggesting that emotion concept formation can be explained by the proposed model.
\end{abstract}


\section{Introduction}
Emotions play a central role in human psychology and behavior, and they are deeply connected to all aspects of our daily lives, including decision making, human relationships, and health. However, despite emotions playing such an important role in the lives of people, their formation process remains unclear. In this study, we approached this problem from a constructionist approach, based on constructed emotion theory \cite{barrett2017emotions}, which has exhibited considerable influence on recent emotion-related research.

Constructed emotion theory posits that emotions are formed by integrating information from interoception, which refers to sensations related to the internal environment of the body such as internal organs, and exteroception, which refers to sensations experienced outside the body such as sight, hearing, smell, taste, and touch \cite{damasio2003feelings, barrett2017emotions, smith2019}. Barrett \cite{barrett2017emotions} stated that, ``in every waking moment, your brain uses past experience, organized as concepts, to guide your actions and give your sensations meaning. When the concepts involved are emotion concepts, your brain constructs instances of emotion.'' Researchers have expressed and debated varied perspectives related to emotions; however, most researchers believe that emotion concepts are acquired through experience \cite{smith2019}.

In this case, how are emotion concepts acquired? In this study, based on constructed emotion theory, we considered emotion concepts to be categories  formed from interoceptive and exteroceptive information associated with specific emotions. When people receive a stimulus, they form a basic emotional state called a core affect based on the interoception of the stimulus \cite{russell1999core, russell2003core, moriguchi2013neuroimaging}. Then, the brain integrates core affect and exteroceptive information, and further categorizes the integrated sensory information through emotion concepts acquired from past experience \cite{barrett2017emotions, barrett2015interoceptive, smith2019, moriguchi2013neuroimaging}. Instances of specific emotions such as sadness and anger are generated from the concepts formed in this process. Here, emotion concepts do not refer to independent categories with clear boundaries. To begin with, emotion concepts are not fixed and are constantly changing. When the brain receives certain sensory information, it attempts to identify the cause. Particularly, the brain attempts to identify a reasonable combination of past sensory information (past experiences) that resulted in an emotion concept to identify where the specifically acquired information was generated. Thus, emotion concepts are stochastically composed of multiple dynamically changing categories \cite{barrett2017emotions, barrett10.1093/scan/nsw154, clark2017multivoxel, wilson2011grounding}.

\IEEEpubidadjcol
Why must people form emotion concepts? We present an encounter with a dog at the side of a street as an example. When a person sees a dog, the person generates a certain type of physiological reaction, such as an increase in heart rate. A core affect is formed from this interoceptive information, which is integrated with visual and auditory information that is obtained related to the dog, such as the facial expression, size, and barking action of the dog. Categorization is then performed by matching the core affect with pre-existing emotion concepts, which have been formed by past experiences. Therefore, if a person was bitten by a similar dog in the past, then the emotion of ``fear'' is generated by the emotion concept that reflects that experience. Simultaneously, the person predicts tactile information such as ``pain'' and visual information such as ``red blood'' that are associated with the emotion of ``fear'' and attempts to perform evasive actions. Therefore, emotion concepts not only assign meaning to sensory information and generate emotional instances, but also aid in predicting unobserved information from the acquired information and prescribing behavior \cite{barrett2017emotions, barrett10.1093/scan/nsw154}. This formation of categories for predicting unobserved information is an important function of emotion concepts and is one reason why people form emotion concepts.

However, not all people experience the same emotion in response to the same stimulus. In the above example, if the observed dog is similar to a dog owned by the person, then they would probably not experience the emotion of ``fear.'' This is because, even if people have the same core affect state, their past experiences (happy memories of playing with their own dog) differ, resulting in the categorization of different emotion concepts. The results of this categorization update the existing emotion concept. The influence of this type of top-down processing based on emotion concepts produces the individual differences in emotion in response to the same external stimulus.

As explained previously, interoception and exteroception are integrated to form an emotion concept, and the brain generates emotions and predicts unobserved information through this concept. However, little progress has been made in examining the types of information processes that actually occur when the interoceptive and exteroceptive information are integrated.
Research has also been conducted in the field of symbol emergence in robotics to form concepts using multimodal categorization for examining the concept formation process exhibited by people \cite{Nakamura, hagiwara, miyazawa2019integrated, attamimi2014integration, fadlil2013integrated, araki2012online}. Multimodal categorization in symbol-emergent robotics is the idea of learning categories that are similar to human sensations from multimodal information without supervision \cite{Nakamura}. The learned categories correspond to concepts, and the prediction of unobserved information based on concepts is considered to be understanding. Concept formation using multimodal categorization has previously focused on externally observable aspects such as objects and places. Nakamura et al. \cite{Nakamura} constructed a robot with sensors that can acquire visual, auditory, and tactile information, and they showed that object concepts could be acquired by categorizing this information. Moreover, Hagiwara et al. \cite{hagiwara} showed that a robot can predict location names and categories in a manner similar to human predictions based on vision, position, and word information. Miyazawa et al. \cite{miyazawa2019integrated} used multilayered multimodal latent Dirichlet allocation (mMLDA) \cite{attamimi2014integration} to propose an integrated cognitive architecture for robots to simultaneously learn behavior and language, and they demonstrated the effectiveness of the model through experiments using an actual robot. However, human mental states such as emotions have barely been addressed in this field. Ohmori et al. \cite{ohmori} conducted multimodal concept formation using physiological signals based on the eating task performed by humans as an approach to cognitive structures including emotions. However, the number of stimuli and subjects in this study were limited, and the data may not have been sufficiently controlled.

Therefore, in this study, we applied multimodal categorization to emotion concept formation and attempted to model it. Particularly, we constructed a model using mMLDA \cite{attamimi2014integration}, which is a probabilistic generative model. We then trained a model for each subject using multimodal data obtained when emotion-evoking stimuli were presented to people, and we verified the validity of the model by comparing the categories formed by the model with subjective emotional reports. We also used the formed categories to investigate the prediction of unobserved information, which is an important element of the concept. We used the above probabilistic generative model to express emotion concept formation based on constructed emotion theory, 
and we believe this constructive approach toward emotions may yield a clue to the elucidation of human emotions. 

\section{Emotion concept formation model}
Here, we describe multimodal latent Dirichlet allocation (MLDA), which is the basis of the mMLDA. Subsequently, we provide an overview of the model used in this research and its implementation method.

\subsection{Multimodal latent Dirichlet allocation}
\label{subsection:MLDA}

\begin{figure}[t]
\centering
\includegraphics[width=0.8\hsize]{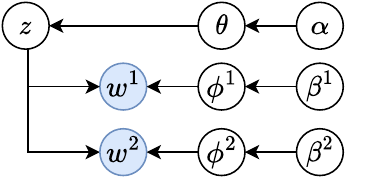}
\caption{Graphical model of multimodal latent Dirichlet allocation. }
\label{fig:MLDA}
\end{figure}

MLDA \cite{Nakamura} is a generative model based on latent Dirichlet allocation (LDA) \cite{LDA_1} that can form concepts from multimodal information. Here, a ``concept'' refers to a category obtained by grouping different types of inputs. Fig. \ref{fig:MLDA} shows the graphical model of MLDA when the number of modality information is two. In the figure, \(z\) represents the conceptual category of observed information \(w^* \in \{w^{1}, w^{2}\}\), which is generated from a multinomial distribution with parameters \(\theta\), \(\phi^* \in \{\phi^{1}, \phi^{2}\}\), respectively. They are expressed using the following equations:

\begin{align}
    z &\sim Mult(\theta), \\
    w^* &\sim Mult(\phi^*).
\end{align}

\noindent Moreover, \(\alpha\), \(\beta^* \in \{\beta^{1}, \beta^{2}\}\) represent the hyperparameters of the Dirichlet distribution that generate \(\theta\) and \(\phi^*\), respectively, which can be expressed as follows:

\begin{align}
    \theta &\sim Dir(\alpha), \\
    \phi^* &\sim Dir(\beta^*).
\end{align}

\noindent The categorizations in MLDA involve estimating a concept \(z\) from the obtained observed information \(w^*\) and optimizing it through parameters \(\theta\) and \(\phi^*\). By repeatedly updating parameters for the observed information, past experiences can be accumulated as new concepts. Gibbs sampling \cite{Gibbs} was used for this estimation. The estimation of concept \(z\) using Gibbs sampling is expressed by the following equation \cite{miyazawa2019integrated}:

\begin{align}
    &P(z_{mij} = k | \bm{W}, \bm{Z}^{\backslash{mij}}, \alpha, \beta^m) \nonumber \\
    &\propto (n_{k,j}^{\backslash{mij}} + \alpha) \left( \frac{n_{m, w^{m}, k}^{\backslash{mij}} + \beta^m}{n_{m, k}^{\backslash{mij}} + W^m \beta^m} \right).
\end{align}

\noindent Here, \(\bm{W}\) is a set of modality information, \(W^m\) denotes the number of dimensions of the \(m\)th modality information, and \(\backslash\) indicates the exclusion of information. Moreover, \(n_{m, w^m, k, j}\)  denotes the number of times that category \(k\) is assigned to feature \(w^m\) of \(m\)th modality information in the \(j\)th observed information,
and \(\bm{Z}^{\backslash{mij}}\) is the set of all concepts excluding the concept category \(z_{mij}\) that is assigned to the \(i\)th information of modality \(m\) in the \(j\)th observed information. If \(j\) is not added, it means the number of occurrences in all data, not the \(j\)th data. Furthermore, \(n_{k,j}^{\backslash{mij}}\), \(n_{m, w^{m}, k}^{\backslash{mij}}\), and \(n_{m, k}^{\backslash{mij}}\) denote the number of times category \(k\) is assigned among the \(j\)th observed information, excluding the information that is currently considered, number of times 
that category \(k\) is assigned to feature \(w^m\) of modality \(m\)
, and total number of categories \(k\) assigned with modality \(m\), respectively. This is repeated until \(n_*\) converges; moreover, the values at the time of convergence are used to find \(\theta\) and \(\phi^*\) using the following equations:

\begin{align}
    \theta_{kj} &= \frac{n_{k,j} + \alpha}{n_j + K\alpha}, \\
    \phi^{m}_{w^{m},k} &= \frac{n_{m,w^{m},k} + \beta^m}{n_{m,k} + W^m \beta^m},
\end{align}

\noindent where \(K\) denotes the total number of categories.

Moreover, the learned model can be used to predict the categories of the unobserved data. If the observed information of each modality of actually observed data is \(\bm{w}_{obs}\), then the concept \(\hat{z}\) can be estimated as follows:

\begin{align}
\label{MLDA:new_category}
    \hat{z} \sim P(z | \bm{w}_{obs}) = \int P(z|\theta)P(\theta|\bm{w}_{obs})d\theta.
\end{align}

\noindent The category that maximizes this \(\hat{z}\) is the category of the actually observed data. This is expressed as follows:

\begin{align}
    k = \underset{z}{\mathrm{argmax}} \ P(z | \bm{w}_{obs}).
\end{align}

\noindent This can also be used to predict unobserved modality information. The unobserved modality \(w\) is predicted by the following equation:

\begin{align}
\label{MLDA:prediction}
    P(w | \bm{w}_{obs}) &= P(w | z) P(z | \bm{w}_{obs}) \\ \nonumber
    &= \!\! \int \!\! \sum_{z}  \!\!  P(w | z) P(z|\theta)P(\theta|\bm{w}_{obs})d\theta,
\end{align}

\noindent where \(P(\theta|\bm{w}_{obs})\) in Eqs. (\ref{MLDA:new_category}) and (\ref{MLDA:prediction}) can be obtained by determining \(\theta\) using the aforementioned Gibbs sampling. 

Thus, MLDA can be used to form concepts from unsupervised multimodal information. However, MLDA can only handle a single concept and cannot express relationships between concepts. Therefore, in this study, we constructed an mMLDA model, which is an extension of MLDA.

\subsection{Proposed model}

\begin{figure*}[t]
\centering
\includegraphics[width=\hsize]{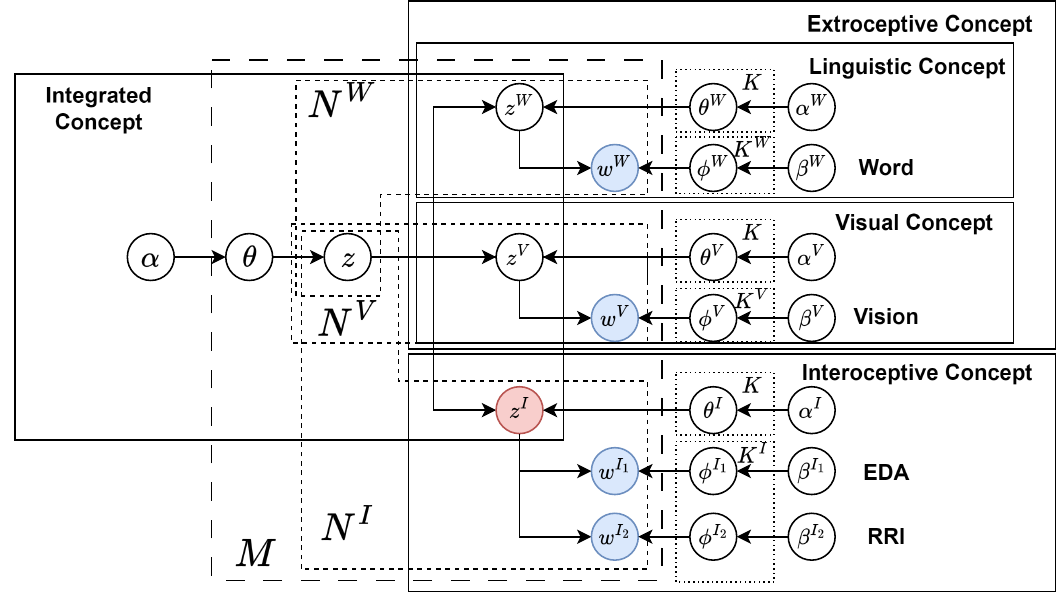}
\caption{Graphical illustration of emotion concept formation model.}
\label{fig:mMLDA}
\end{figure*}

mMLDA \cite{attamimi2014integration} is a generative model that probabilistically expresses relationships between multiple concepts by connecting LDA and the above-mentioned MLDA in a hierarchical manner. Multiple LDAs and MLDAs can be prepared as lower layers corresponding to each piece of information, and MLDA can be arranged as an upper layer to integrate them to efficiently express the categories (i.e., relationships between concepts). In this study, we set up interoception and exteroception modules as the lower layers. Human exteroception and interoception are inherently diverse; however, in this study, we used vision and word information as exteroception and physiology information such as electrodermal activity and heartbeat waveforms as interoception.

Fig. \ref{fig:mMLDA} shows a graphical model of the emotion concept formation model that was constructed based on the aforementioned ideas. In the figure, \(z^* \in \{z^{I}, z^{V}, z^{W}\}\), where \(z^{I}\), \(z^{V}\), and \(z^{W}\) represent the concepts of interoception, vision, and word, respectively; \(z\) represents the integrated concept that captures the relationships among these concepts. These are generated from a multinomial distribution with parameters \(\theta^* \in \{\theta^{I}, \theta^{V}, \theta^{W}\}\) and \(\theta\). Moreover, \(w^* \in \{w^{I_1}, w^{I_2}, w^{V}, w^{W}\}\) is the observed information related to physiology (\(w^{I_1}\) and \(w^{I_2}\)), vision, and words, respectively. These are generated from a multinomial distribution with \(\phi^* \in \{\phi^{I_1}, \phi^{I_2}, \phi^{V}, \phi^{W}\}\).  Similar to MLDA, categorization involves estimating the parameters \(\theta\), \(\theta^*\), and \(\phi^*\) from the observed information \(w^*\), and the Gibbs sampling was used for estimation.



By using mMLDA, unobserved information that spans concepts can also be predicted. 
In Fig. \ref{fig:mMLDA}, \(\alpha\), \(\alpha^* \in \{\alpha^{I}, \alpha^{V}, \alpha^{W}\}\), and \(\beta^* \in \{\beta^{I_1}, \beta^{I_2}, \beta^{V}, \beta^{W}\}\) represent the hyperparameters of the Dirichlet distribution that generate \(\theta\), \(\theta^*\), and \(\phi^*\), respectively. Furthermore, \(M\) denotes the total number of data, \(N^* \in \{N^{I}, N^{V}, N^{W}\}\) is the total amount of information in each modality, and \(K\) and \(K^* \in \{K^{I}, K^{V}, K^{W}\}\) are the number of categories for each concept, respectively. In this study, the hyperparameters \(\alpha\), \(\alpha^*\), and \(\beta^*\) were set to 1.0.

In this study, we assumed that a emotion concept is represented in the interoception category \(z^I\), which is influenced by exteroception. Core affect, which is the perception of interoception, is said to be expressed on a plane consisting of two axes (valence and arousal) \cite{russell2003core}. Emotion concept formation is speculated to occur with this core affect as a basis and adding exteroceptive information to it. Therefore, in this model, the emotion concept is expressed in \(z^I\), which is a core affect influenced by exteroceptive information.


\begin{figure*}[t]
\centering
\includegraphics[width=\hsize]{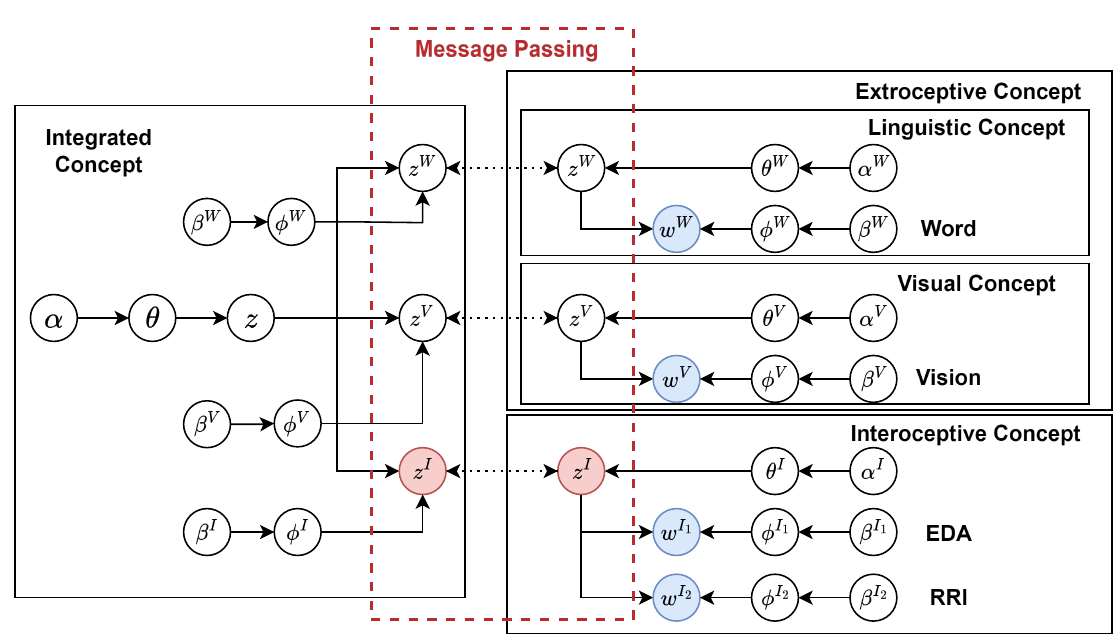}
\caption{Overview of emotion concept formation model implemented with Serket.}
\vspace{-2mm}
\label{fig:Serket}
\end{figure*}

Serket \cite{nakamura2018serket} was used for model implementation. Serket is a framework that enables the construction of cognitive models by connecting modules while each module maintains its programmatic independence. Serket can achieve inference with mMLDA based on multiple LDA and MLDA inferences and the message passing approach. Parameter updates are based on a single LDA or MLDA inference; therefore, the learning and prediction methods described in Section \ref{subsection:MLDA} can be used. Fig. \ref{fig:Serket} shows an overview of the model illustrated in Fig. \ref{fig:mMLDA} when it is implemented using Serket. 
First, the probabilities shown below can be computed by Gibbs sampling.



\begin{align}
\label{Serket:message}
    \hat{z}^{I}_{j} \sim P(z^{I}_{j} | w^{I_1}_j, w^{I_2}_j), \\
    \hat{z}^{V}_{j} \sim P(z^{V}_{j} | w^{V}_j), \\
    \hat{z}^{W}_{j} \sim P(z^{W}_{j} | w^{W}_j).
\end{align}

\noindent Here, \(w^*_j\) is the observed information of the \(j\)th data, \(z^*_j\) is the latent variables of the \(j\)th data.
Next, the values of the shared latent variables are inferred stochastically using a integrated concept model:

\begin{align}
\label{Serket:message2}
    P(z^{C} | \hat{z}^{I}_{j}, \hat{z}^{V}_{j},  \hat{z}^{W}_{j})\\ \nonumber
    &= \!\!  \sum_{z} \!\!  P(z^{C} | z) P(z|\hat{z}^{I}_{j}, \hat{z}^{V}_{j},  \hat{z}^{W}_{j}).
\end{align}

\noindent Here, \(C \in \{I, V, W\}\) is the concept.
These probabilities are also represented by finite and discrete parameters, which can be communicated using the message passing approach.
These parameters are sent to each concept model, where the latent variables assigned to the modality information of the concept are determined using Gibbs sampling.

\begin{align}
\label{Serket:message3}
    z^{C}_{mij} \sim P(z^{C} | \bm{W}, \bm{Z}^{\backslash{mij}},\alpha, \beta^m) P(z^{C} | \hat{z}^{I}_{j}, \hat{z}^{V}_{j},  \hat{z}^{W}_{j} ),
\end{align}

\noindent where \(\bm{W}\) represents all the information, and \(\bm{Z}^{\backslash{mij}}\) represents a set of latent variables, except for the latent variable assigned to \(i\)th information of modality \(m\) of the \(j\)th observation. 
All the latent variables were learned in a complementary manner.



\section{Experiment}
\label{section:実験}
The experiment was divided into three phases: data collection, model learning, and validation. In the data collection phase, we obtained information on each modality from the subjects through a visual stimulation task. In the model learning phase, we preprocessed the information related to each modality, which was obtained in the subject experiment for model learning. In the validation phase, \(z^I\) of the learned model was compared with the subjective emotional reports of the subjects. We also attempted to predict unobserved information in other modalities from observed information in a specific modality for unknown images.

\subsection{Data collection}
A subject-based experiment was conducted to collect data for model learning. Thirty-one subjects (14 males and 17 females with ages ranging from 20 to 40 years) were selected for the experiment. Sixty images that aroused emotion were extracted from the International Affective Picture System (IAPS) \cite{IAPS} and presented to the subjects. The physiological signals were obtained using an E4 wristband \cite{e4wristband} and a myBeat WHS-1 wearable heart-rate sensor to obtain the electrodermal activity (EDA; 4 Hz) and heartbeat waveform (128 Hz), respectively.  These data can be obtained along with time stamps using a designated dongle connected via Bluetooth. myBeat can measure the electrical activity of the heart at 128 Hz by being attached to the electrode pad developed for myBeat and attaching it near the heart. The data were obtained with a timestamp by the attached receiver for the repeated cycles of the experiment. Subsequently, the subjects were asked to complete an emotional self-evaluation scale to understand their emotional state regarding the images. IAPS publishes images as well as values of Self-Assessment Manikin (SAM) evaluation \cite{SAM}. SAM is an emotional self-evaluation scale that uses a nine-point scale with illustrations for each of the three evaluation axes: valence, arousal, and dominance. The 60 images used in this experiment were selected so that this value was varied. 

Note that the physiological signals are acquired as information on interoceptors, which are the interoception sources, and we selected the physiological signals that are most often used in research and that are relatively less burdensome to subjects \cite{leiner2012eda, van1993heart}. We also used Google speech recognition \cite{rajput2017comparing} for obtaining word information, with the subjects being asked to provide voice input in Japanese regarding their feelings about the presented images. 
Then, subjects were asked to input individual words as much as possible.

Moreover, SAM was used as an emotional self-evaluation scale for images. In the experiment, we obtained all values for valence, arousal, and dominance; however, dominance is difficult to teach, and subjects often chose it incorrectly. An experiment by Lang et al., who created SAM, showed that the correlation coefficients between pleasant and aroused emotions and the SAM valence and arousal values, respectively, were 0.96 and 0.95, respectively, whereas that between a dominant emotion and the SAM dominance value was 0.23, which is relatively low \cite{SAM}. Lang et al. posited that this was due to the fact that the dominance value inherently represents the personal sense of domination of the subject in a particular situation; however, subjects often interpret this as a sense of domination of the object in the image. In practice, many similar cases were observed in the present experiment as well; consequently, the dominance value was not used in the analysis, and only the valence and arousal values were used.

The durations of the experiments were set as follows: 75 s for one cycle, 6 s for image display, 50 s for word information input, 15 s for SAM response, and 4 seconds in total for the waiting time between each phase. A three-minute break was also included between each session. Moreover, before the start of the experiments, we conducted five minutes of resting state measurements and a practice session for the subjects to become used to basic operations. After the completion of the experiments, subjects were asked to complete a questionnaire regarding their personal characteristics. The experiment time, including these times, was 2.5–3 h per person. Of all the subjects in this study, 29 (13 males, 16 females) who did not have major physiological signal deficits were included in the analysis. Furthermore, for subjects who showed deficits in only some of their physiological signals, all other data excluding the deficient data were included in the analysis.

\subsection{Model learning}
The model shown in Fig. \ref{fig:mMLDA} was trained using the observed information that was obtained by converting the modality information obtained through experiments into vector representations. The information for each modality was processed as explained subsequently.

\subsubsection{Physiological information}
We used 6 s of physiological signals during image presentation as modality information for the interoception module. For data preprocessing, we applied a high-pass filter that cuts frequencies below 0.05 Hz of the EDA, and the data was then smoothed using a moving average over a one-second period. Standardization was also conducted using experimental data. For the heartbeat waveform, the R-R interval (RRI) was calculated using ``findpeaks,'' which is a peak detection algorithm in MATLAB from MathWorks. However, the RRI obtained using this method is not equally spaced; therefore, it was converted to equal intervals using the Akima algorithm for one-dimensional interpolation  (sampling interval: 128 Hz) \cite{akima1970new} and standardized in the same manner as the EDA data. Furthermore, each preprocessed data was compressed using a vector-quantized variational autoencoder (VQ-VAE) \cite{van2017neural} and converted into features. VQ-VAE is a model that extends the variational autoencoder and can express latent variables in a discrete manner. In this study, each model was trained in advance using the physiological signals that were obtained during rest and practice. Each dataset was represented by a 11,383-dimensional latent variable using an encoder, and a discrete representation of the latent variable was obtained by comparing it with the 128 codebooks of these pretrained models. The frequency information of this discrete representation was used as observation information for mMLDA. The mean reconstruction errors for the EDA and RRI  data after preprocessing were 0.048 and 0.0033, respectively.

\subsubsection{Vision information}
For the vision information, we used a 1536-dimensional vector obtained by converting the 60 IAPS images used in the experiment into features using Inception-Resnet-v2 \cite{szegedy2017inception}. Inception-Resnet-v2 is a convolutional neural network  that has achieved high performance in image recognition and feature extraction tasks. In this study, we used the pretrained model published by Szgedy et al. \cite{szegedy2017inception} and used the intermediate layer when the 60 images used in the experiment as input were used as observed information.

\subsubsection{Word information}
For the word information, we used a bag-of-words representation of the content entered by the subject during the 50 s of word information input. A bag-of-words representation is a vectorization of information related to frequency of occurrence of words, and the resulting 3626-dimensional frequency information was used as word observation.

\subsubsection{Determination of number of categories}
The number of LDA categories had to be manually determined in advance. The mMLDA used in this study is an extended model of LDA; therefore, the category \(K\) of the integrated concept and category \(K^*\) of each concept had to be determined in advance.

The number of categories \(K^I\) for MLDA, which is the interoception category, was set to four. Various models have been proposed to express emotions \cite{ekman1992argument, tomkins2008affect, russell2003core, bakker2014pleasure, trnka2016modeling}; however, no consensus has been reached on any of them \cite{lindquist2013hundred}. Therefore, in this study, the emotion concept, which is considered to be represented in the interoception category, follows the idea that core affect is expressed on a plane consisting of two axes, valence and arousal \cite{russell2003core}, and we assumed that this category is similar to the space formed by these two axes. Then, we set the number of categories as four, which is the number of quadrants in this space.

The number of LDA categories \(K^V\) and \(K^W\), which are the exteroceptive concepts, is set as 4 and 34, respectively, after training an LDA model with only vision information for each subject, conducting a grid search using the log-likelihood function in the range of 4–60, and considering the mode of all subjects. Moreover, the number of categories \(K\) in MLDA corresponding to an integrated concept was set to 28 after training the MLDA for each subject using the learning results for each concept as observed information, and similarly conducting a grid search using the log-likelihood function in the range of 4–60.

\subsection{Learning method}

\begin{figure*}[t]
\centering
\includegraphics[width=\hsize]{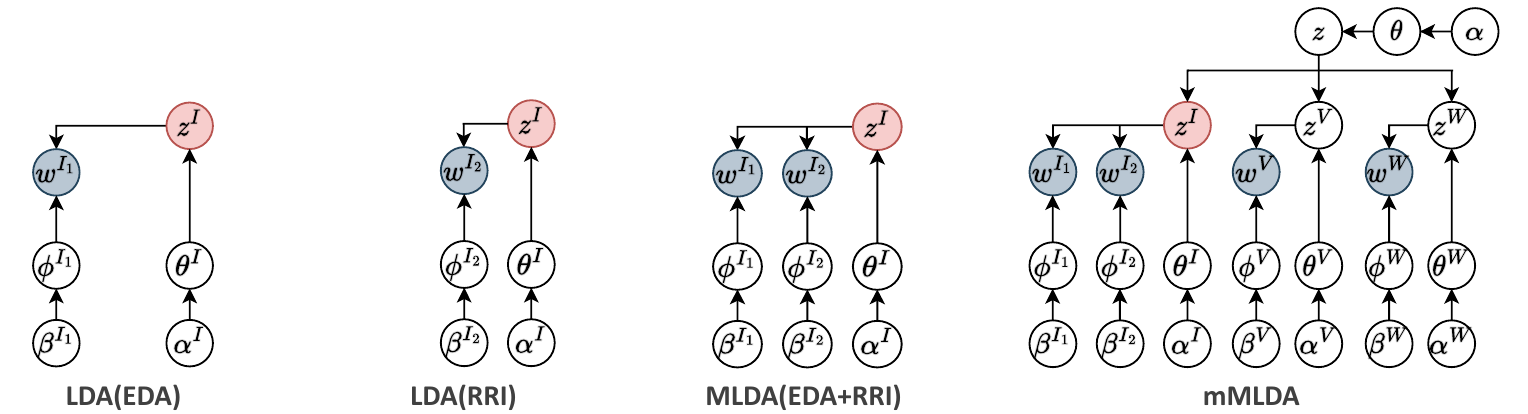}
\caption{Model comparison at the verification stage.}
\vspace{-2mm}
\label{fig:model_comparison}
\end{figure*}

The model was trained using two different methods for comparing the subjective emotional reports of the subjects and predicting unobserved information. To compare the reports, as a comparison target, we prepared the proposed model where some information was intentionally deleted. Fig. \ref{fig:model_comparison} shows four graphical models, including the proposed model (mMLDA). We used the observed information for the 60 images used in the experiment to first train LDA with only EDA, LDA with only RRI, and MLDA with only EDA and RRI, after which the mMLDA model shown in Fig. \ref{fig:mMLDA} was trained. To predict the information, we divided the 60 images used in the experiment into 45 learning data and 15 test data, and we used the learning data to train the mMLDA model. Subsequently, we used the learned model and test data to predict 
each information based on other information. 
This training was conducted for each subject in both cases. Thus, models were trained for the total number of subjects.

\subsection{Verification method}
The model learning results were evaluated from two perspectives: similarity to subjective emotional reports and prediction accuracy of unobserved information.

\subsubsection{Similarity to subjective emotional reports}

\begin{figure}[t]
\centering
\includegraphics[width=\hsize]{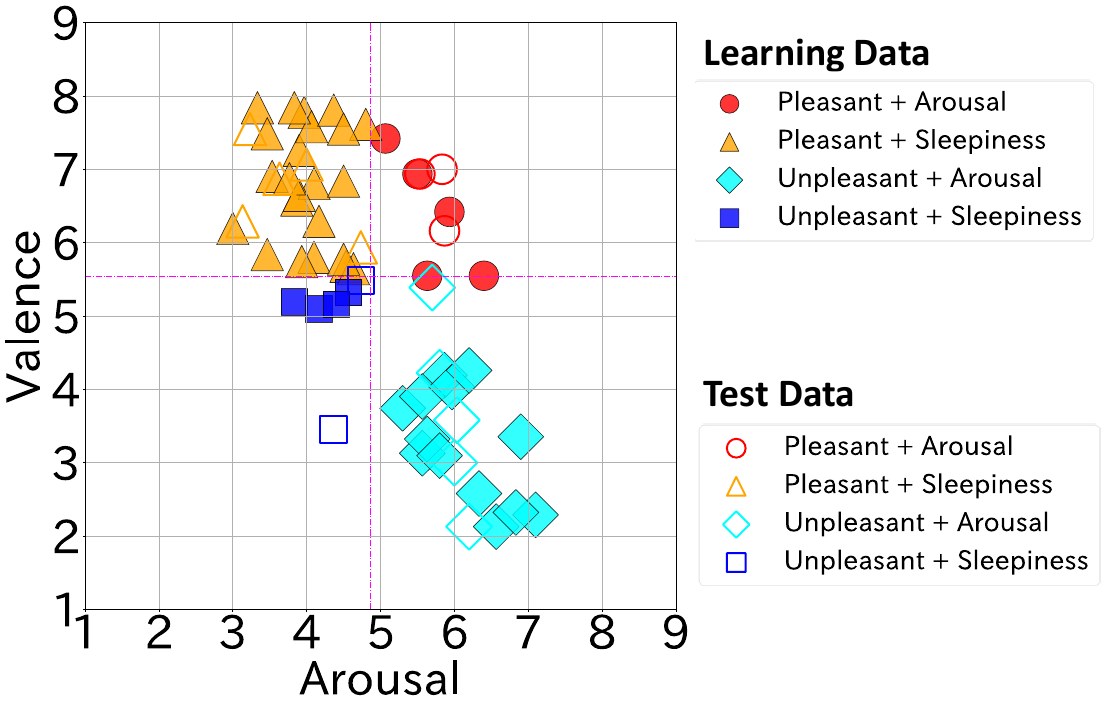}
\caption{Mean values and categories for each image based on the Self-Assessment Manikin evaluation.}
\vspace{-2mm}
\label{fig:SAM}
\end{figure}

We verified whether the interoception categories learned by the model were similar to the subjective emotional reports of people. The SAM responses of the subjects were used for the subjective emotional reports. Fig. \ref{fig:SAM} shows the evaluation of the 60 images used in the experiment, with the vertical and horizontal axes denoting the mean valence and arousal values for all subjects, respectively. This was used as a basis to allocate each quadrant to a category, with the data centroid as the origin. Images were then divided into and classified under the four categories of ``pleasant + arousal,'' ``pleasant + sleepiness,'' ``unpleasant + arousal,'' and ``unpleasant + sleepiness.'' The number of datasets in each category were 8, 26, 18, and 6, respectively. We then calculated the Rand index between the learning categories in the model of each individual and categories labeled based on the SAM results (SAM categories). The Rand index measures the similarity between two clusters, taking a value between 0 and 1, where \(0\) indicates that the two clusters do not match in any pair, and \(1\) indicates that the clusters match completely. The Rand index for each model (LDA, MLDA, mMLDA) was calculated for the number of subjects; therefore, the differences between the models were tested using the Wilcoxon signed-rank test. The significance level \(\alpha\) was set at \(0.05\), and the significance level was corrected by the number of tests using the Bonferroni correction.

We also verified whether emotion concepts were expressed in the categories constructed by mMLDA by comparing the above-mentioned Rand index with chance level. The chance level is the assumption that the four categories were randomly created when calculating the Rand index. The chance level was determined to be \(0.56\).

\subsubsection{Prediction accuracy of unobserved information}
We used the mMLDA model to determine the prediction accuracy. The 60 images used in the experiment were divided into 45 learning data and 15 test data, and the learning data were used for training to predict unobserved modality information based on a set of modality information in the test data. The numbers of learning and test data for the different categories were set to (5, 23, 13, and 4) and (3, 5, 5, and 2), respectively, after considering the overall number ratio. 
We used one set of modality information of test data as the observed information (all other modality information were set as 0).
For the prediction accuracy index, we determined the Kullback--Leibler (KL) divergence between the predicted distribution and the actual distribution.
The predicted distribution means the distribution of the probability of other unobserved modality information obtained by making a prediction from observed information using the learned model.
The actual distribution means the distribution of the actual unobserved modality information normalized 
to sum to 1.

Moreover, emotional words such as ``anger'' and ``joy'' are known to play a particularly important role in concept formation in language \cite{brooks2017role}. Therefore, for word information, we defined words included in the Japanese evaluation polarity dictionary \cite{kobayashi2005} as emotional words, and we used the following two values: the KL divergence between the predicted and actual distributions for all audio-input words, and among them, the KL divergence between the predicted and actual distributions were reconstructed using only the 1,028 emotional words that were actually uttered. For example, when predicting the word information of only emotional words from physiological information, the actual distribution of emotional words can be expressed as \(\bm{w}^{W}_{norm}\), as shown in the following equation.

\begin{align}
\label{prediction_index}
    D_{KL} \left( \bm{w}^{W}_{norm} \parallel P(\bm{w}^{W} | \bm{w}^{I_1}, \bm{w}^{I_2}) \right) \;\;\;\;\;\;\;\;\;\;\;\;\;\;\;\; \nonumber \\
    = \sum \bm{w}^{W}_{norm} \log \frac{\bm{w}^{W}_{norm}}{ P(\bm{w}^{W} | \bm{w}^{I_1}, \bm{w}^{I_2})},
\end{align}

\noindent The KL divergence is an index that measures the difference between two probability distributions, and a smaller value indicates a higher similarity between the distributions. Moreover, we set a chance level in the same manner as when verifying the similarity with subjective emotional reports, and we evaluated the prediction accuracy by examining the occurrence of a significant difference from the chance level. For the chance level, the predicted distribution was assumed to be a uniform distribution, and we used the similarity between the uniform and actual distributions of words calculated using Eq. (\ref{prediction_index}). We also used the paired t-test as the testing method.

\section{Results}

\begin{figure}[!t]
\centering
\subfloat[LDA vs MLDA]{\includegraphics[width=2.5in]{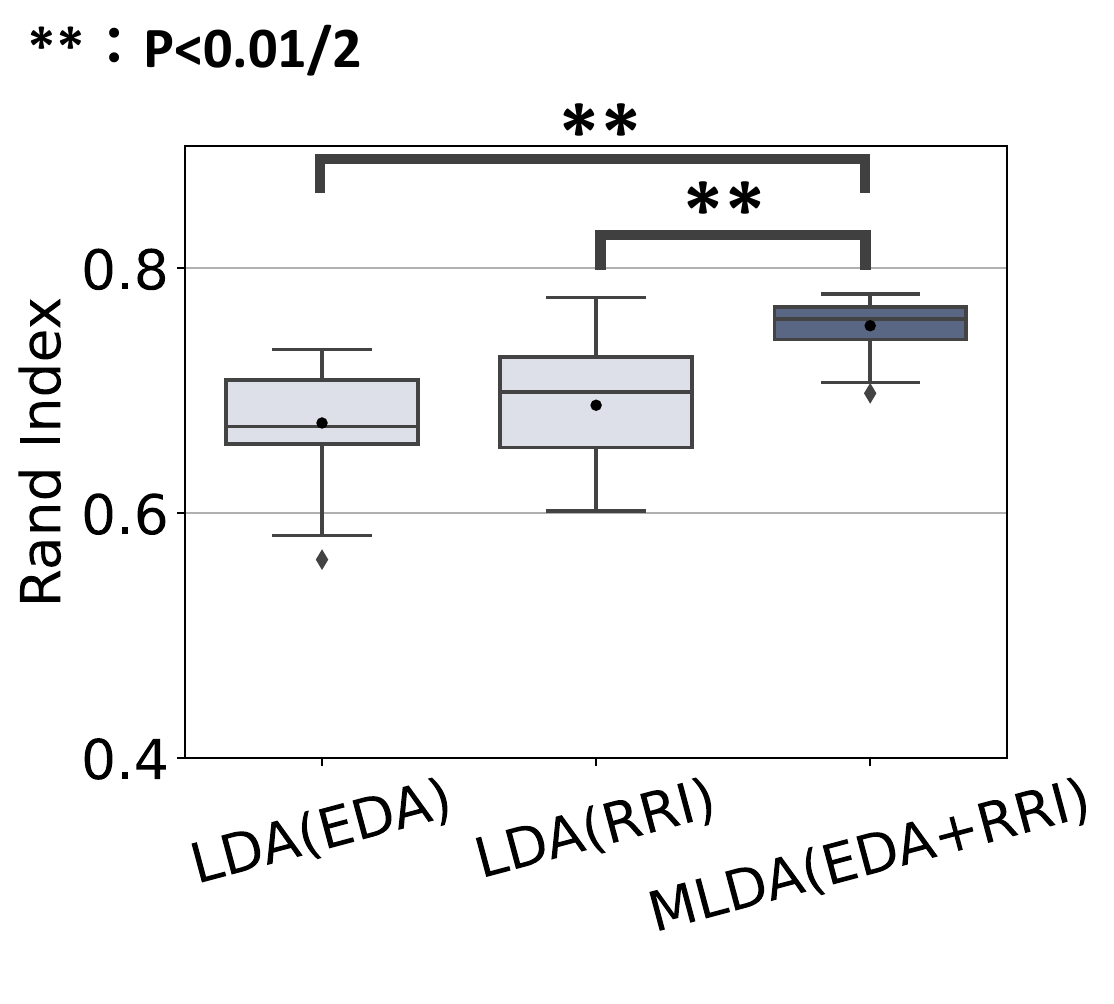}%
\label{fig/rand_index_LDA_vs_MLDA.pdf}}
\hfil
\subfloat[MLDA vs mMLDA]{\includegraphics[width=2.5in]{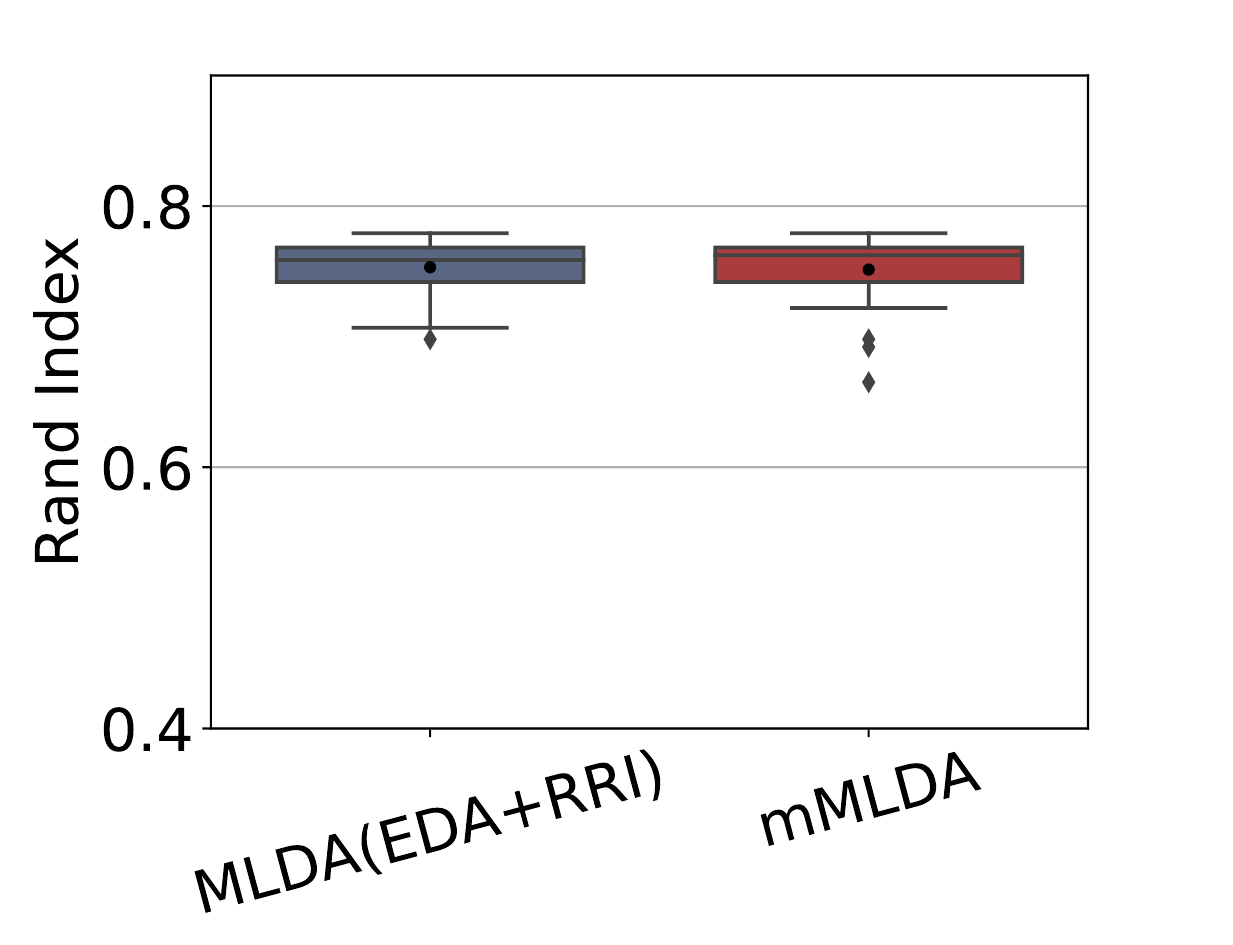}%
\label{fig/rand_index_MLDA_vs_mMLDA.pdf}}
\caption{Rand index of each model.}
\label{fig:rand_index}
\end{figure}


\begin{table}[!t]
\caption{Rand index for each model.}
\label{tab:rand_index}
\centering
\begin{tabular}{c|cc}
\hline
 & Mean & SD\\
\hline
\hline
LDA(EDA) & 0.67 & 0.041\\
\hline
LDA(RRI) & 0.69 & 0.047\\
\hline
MLDA(EDA+RRI) & 0.75 & 0.020\\
\hline
mMLDA(ALL) & 0.75 & 0.027\\
\hline
\end{tabular}
\end{table}


First, for the similarity with subjective emotional reports, Fig. \ref{fig:rand_index} shows the comparison results of the Rand index for each model, and Table \ref{tab:rand_index} lists the mean values and standard deviations. In the comparison between each LDA and MLDA (Fig. \ref{fig:rand_index}(a)), the Rand index of MLDA was significantly higher (vs EDA:\(P=3.7 \times 10^{-9}<0.01/2\), vs RRI:\(P=1.0 \times 10^{-5}<0.01/2\)). This suggests that using multiple sets of physiological information is better as an emotion concept formation. Moreover, a comparison between MLDA and mMLDA did not show a significant difference (\(P=0.73>0.05\)). Thus, in terms of category similarity, the influence of the presence of vision and word information is unclear. Next, we compared the Rand index with the chance level to determine whether the interoceptive sensation categories constructed by mMLDA could be considered as emotion concepts. The mean Rand index of mMLDA was \(0.75\), which was higher than the chance level (\(0.56\)). This suggests that an emotion concept was formed in this study that is more in line with the subjective emotional reports of people than a random emotion concept.

\begin{figure}[!t]
\centering
\includegraphics[width=3.4in]{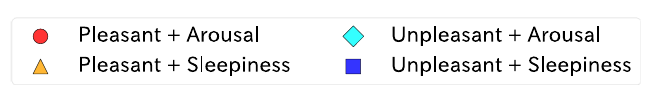}
\subfloat[LDA(EDA)]{\includegraphics[width=1.65in]{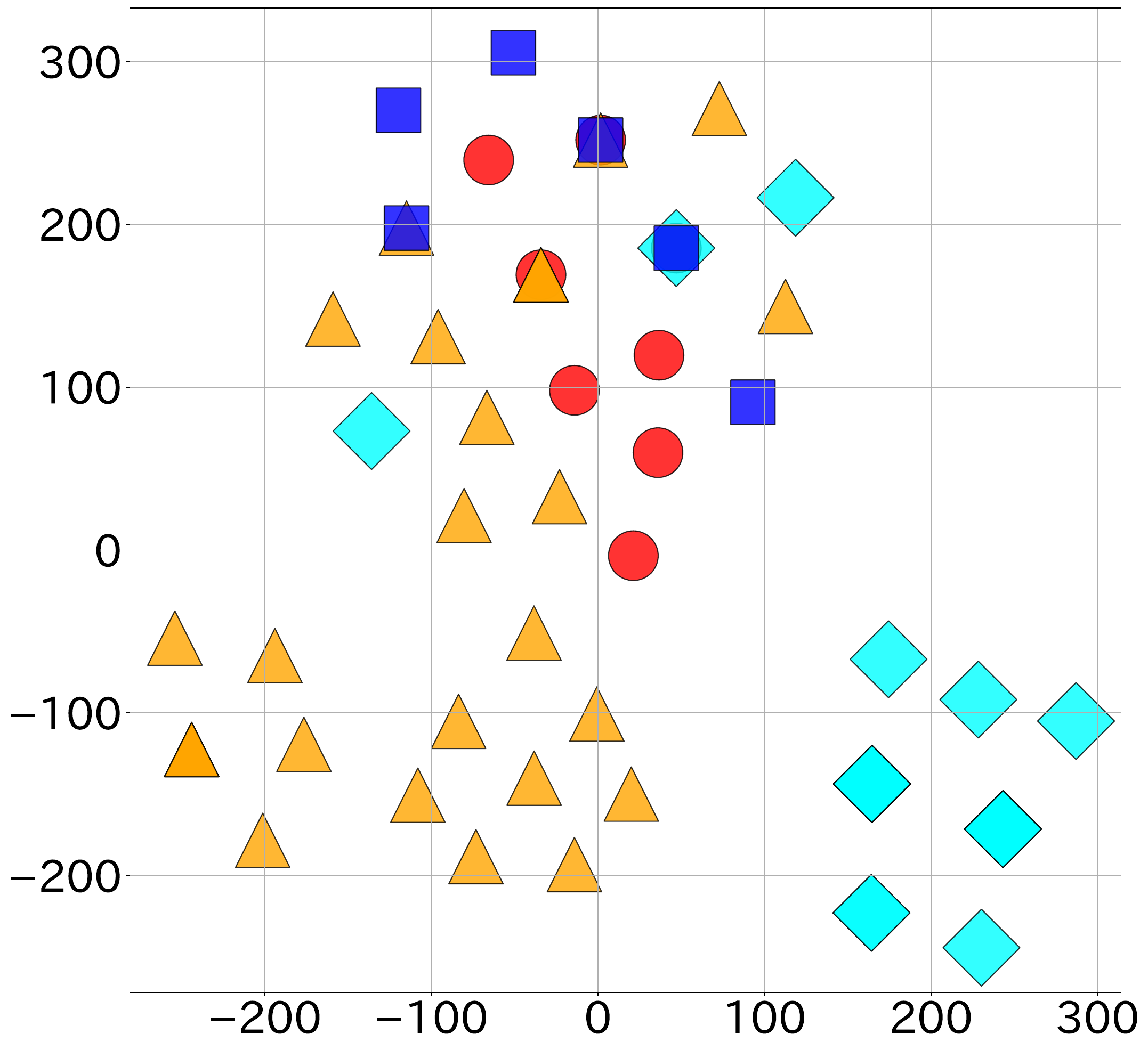}}%
\hfil
\subfloat[LDA(RRI)]{\includegraphics[width=1.6in]{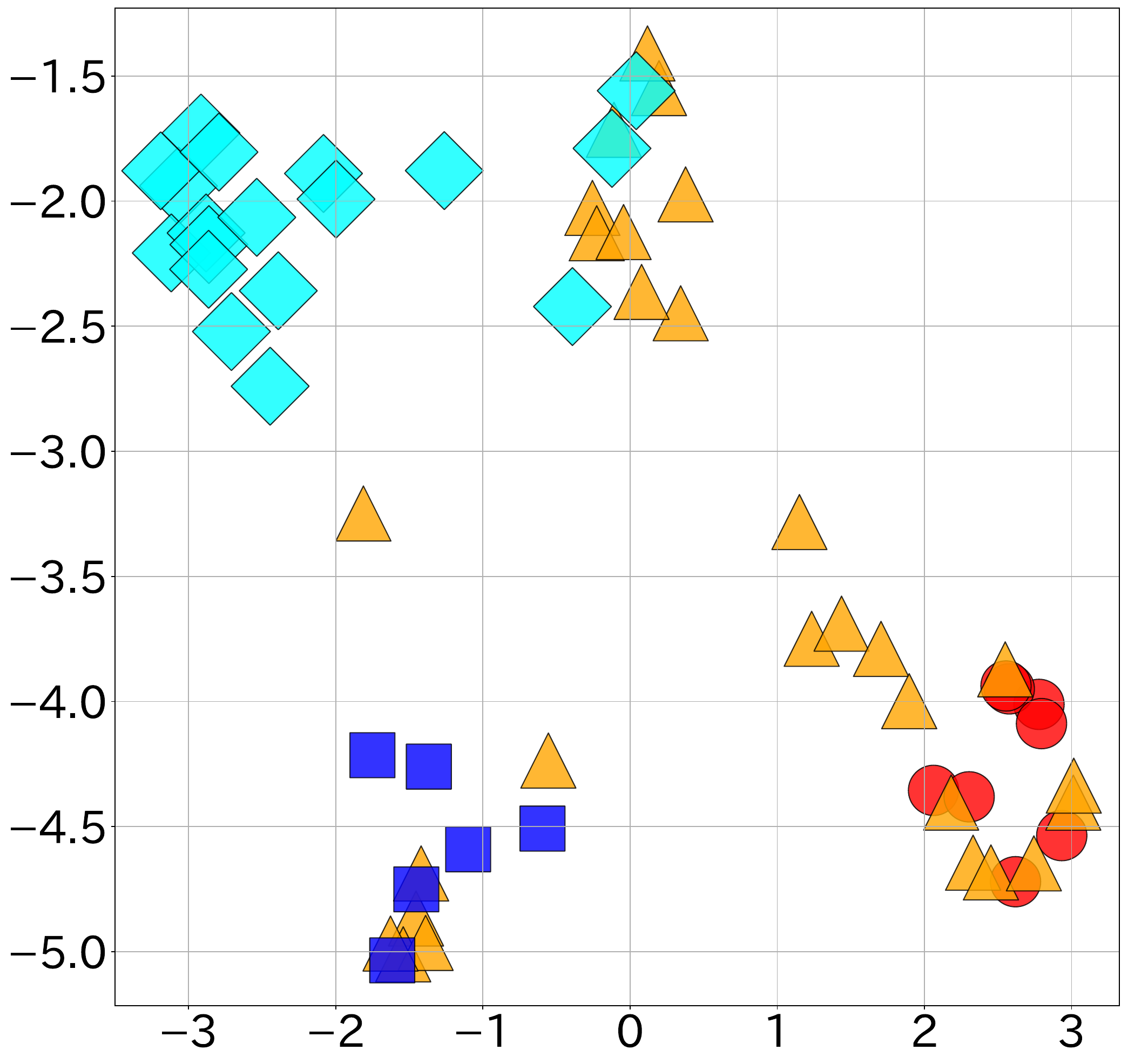}}%
\hfil
\subfloat[MLDA(EDA+RRI)]{\includegraphics[width=1.6in]{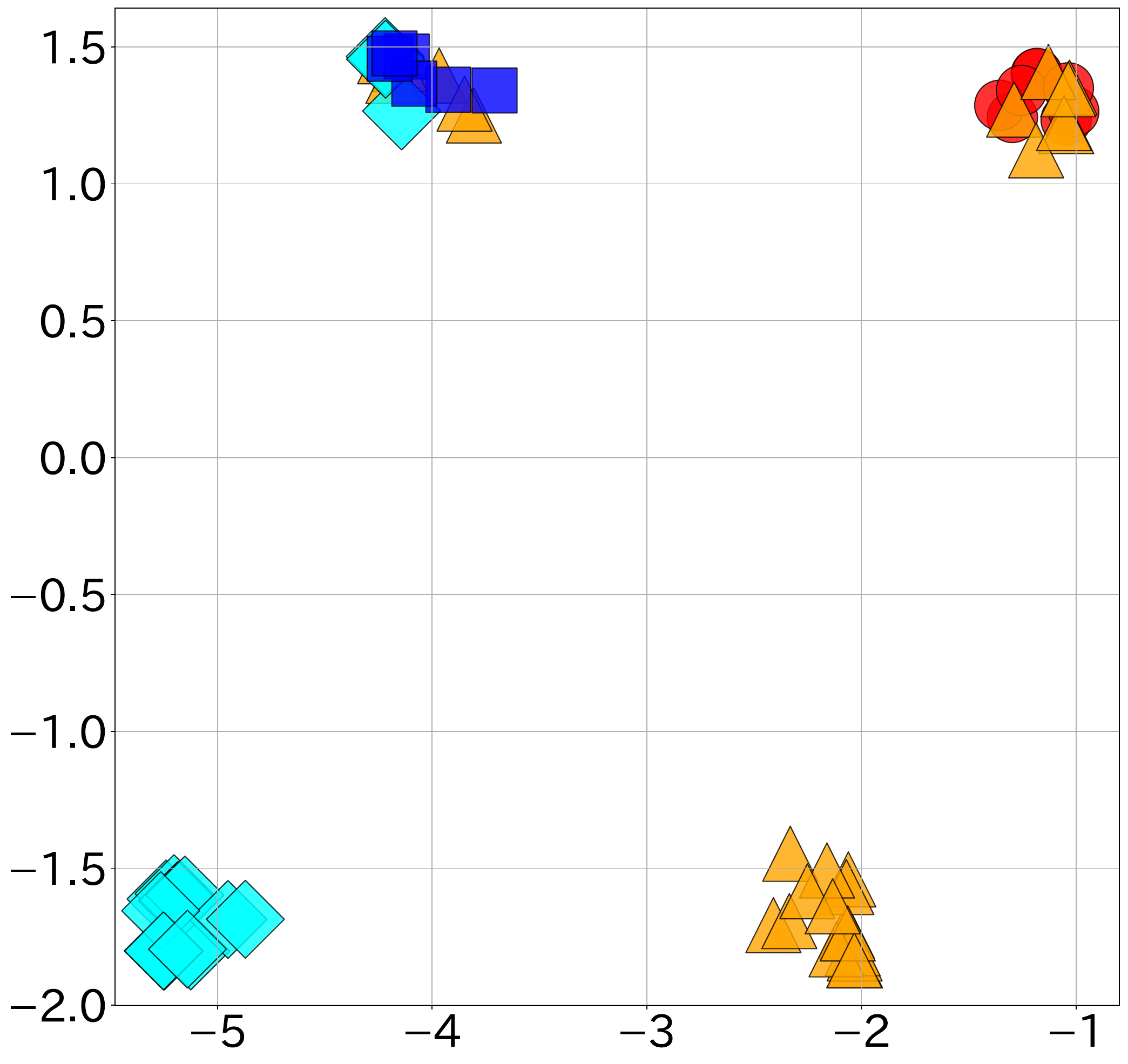}}%
\hfil
\subfloat[mMLDA(ALL)]{\includegraphics[width=1.6in]{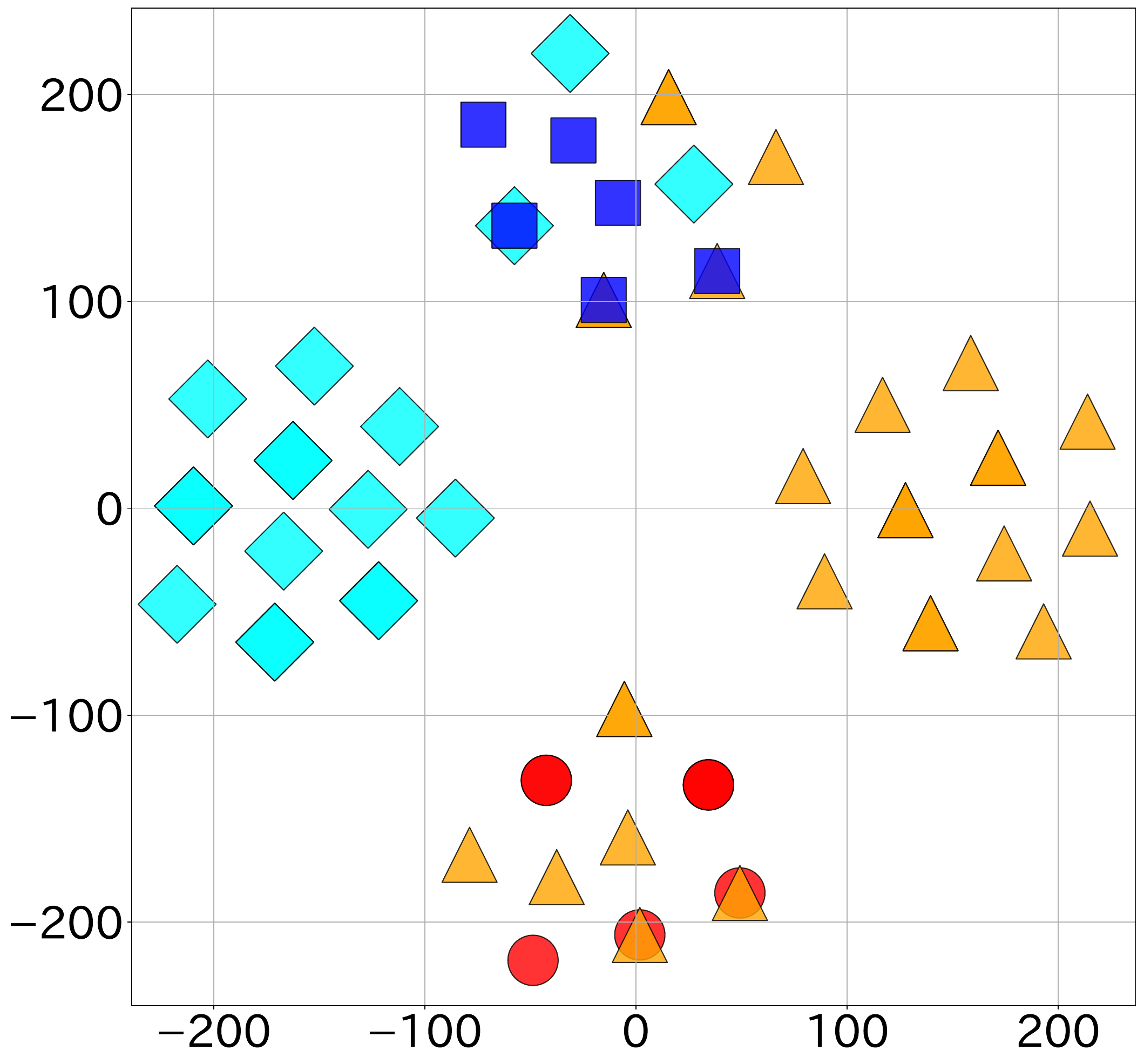}}%
\caption{Concepts of each model visualized using t-distributed Stochastic Neighbor Embedding.}
\label{fig:concept}
\end{figure}


Fig. \ref{fig:concept} shows an example of the emotion concept space of a subject. This refers to a space in which the probability of belonging to each of the four categories, expressed by \(z^I\) in Fig. \ref{fig:mMLDA}, is compressed into two dimensions using t-distributed Stochastic Neighbor Embedding and visualized. As shown in Fig. \ref{fig:concept}(a), the category ``unpleasant + arousal'' can be confirmed in the conceptual space created by LDA (EDA), at the bottom-right side of the figure; however, the results corresponding to the other three categories are mixed together. In contrast, in Fig. \ref{fig:concept}(b), the categories are marginally separated in LDA (RRI); however, the results for ``pleasant + sleepiness'' were mixed with those of the other categories. In Fig. \ref{fig:concept}(c), the four categories are clustered separately for MLDA, and high-accuracy classification was achieved. Furthermore, in Fig. \ref{fig:concept}(d), the categories are clearly separated in mMLDA. Moreover, data for images belonging to ``pleasant,'' ``unpleasant, '' ``arousal,'' and ``sleepiness'' are clustered at the bottom-right, top-left, bottom-left, and top-right sides, respectively, indicating the valence and arousal axes. Qualitative evaluation of the space of 29 participants showed that 10 subjects visually displayed the valence and arousal axes on the conceptual space of mMLDA, which was higher than the 4 subjects for the MLDA.


\begin{figure}[!t]
\centering
\subfloat[EDA]{\includegraphics[width=2.5in]{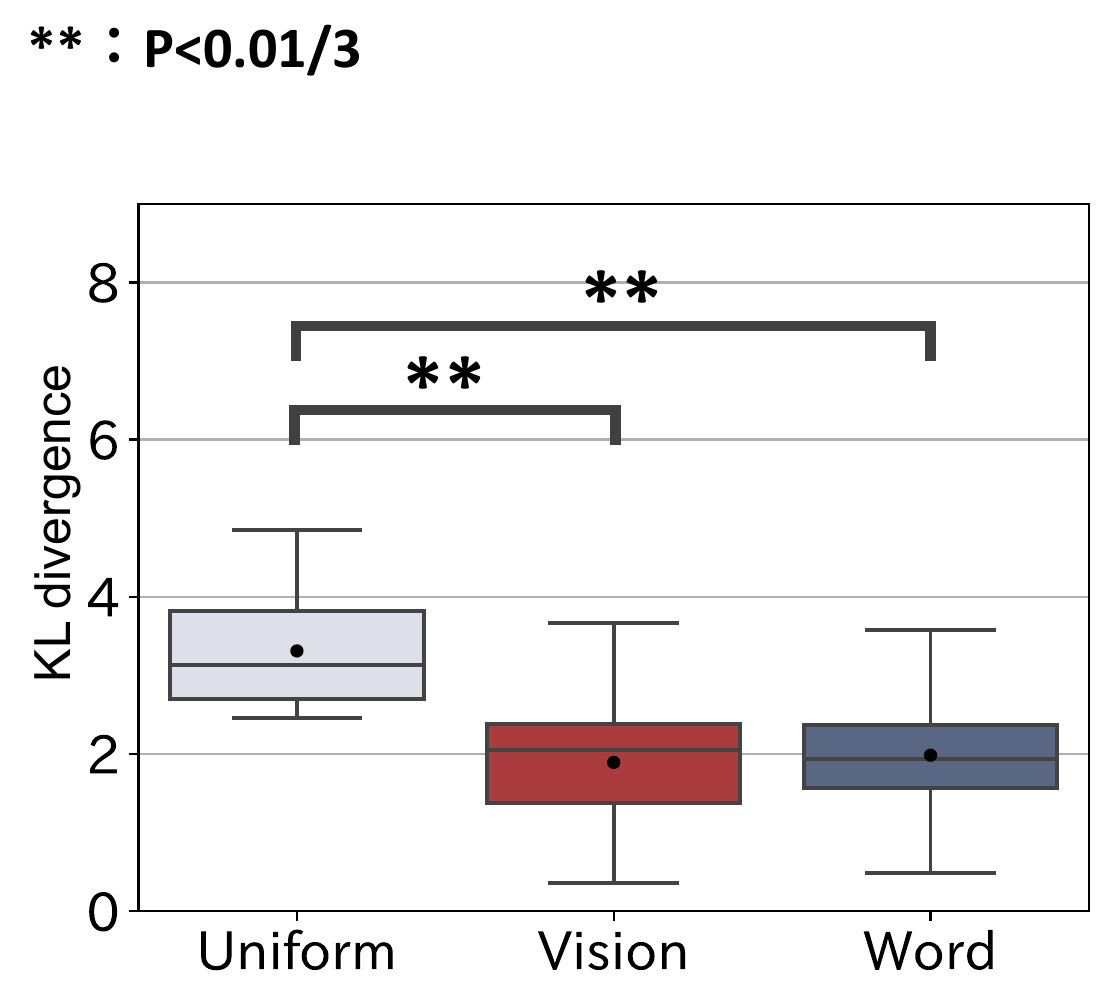}%
\label{fig:KL_to_EDA}}
\hfil
\subfloat[RRI]{\includegraphics[width=2.5in]{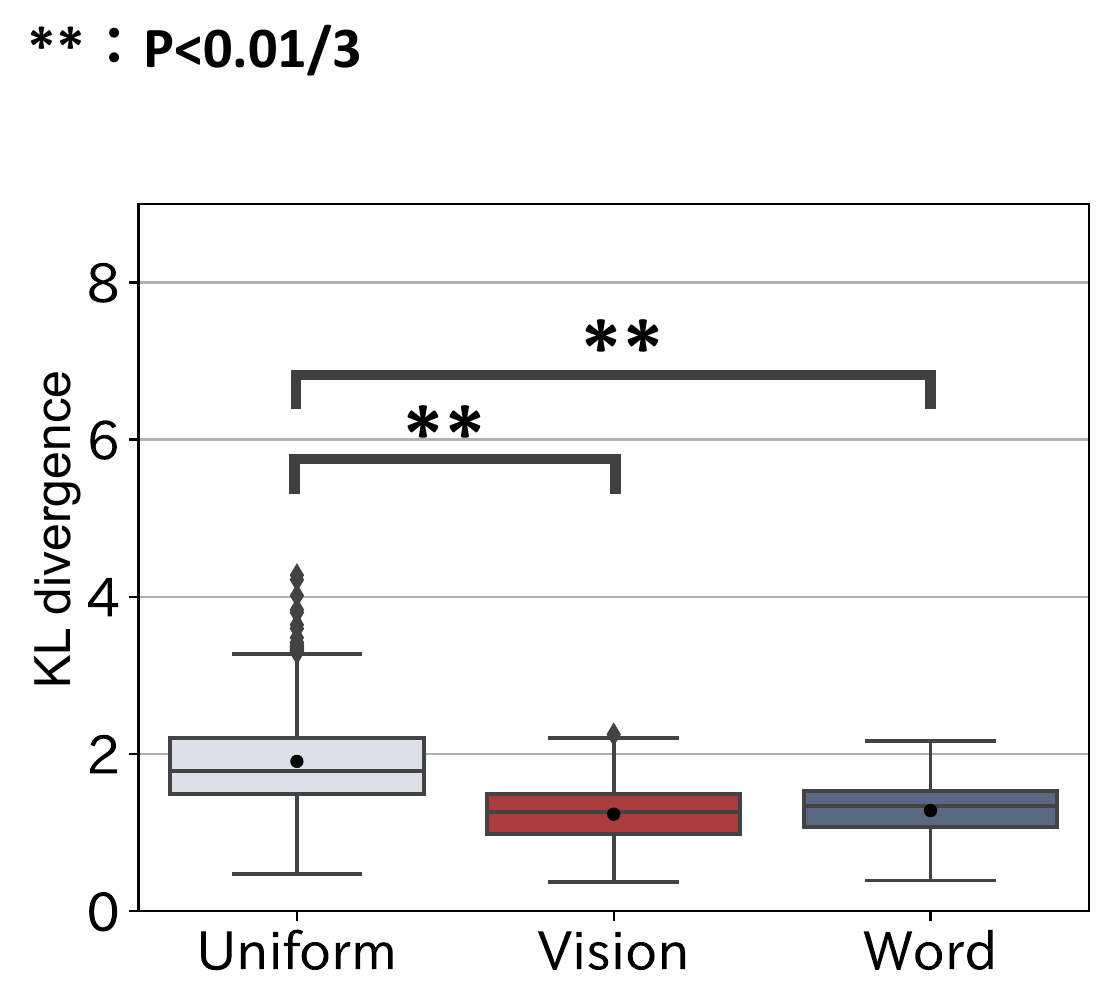}%
\label{fig:KL_to_RRI}}
\caption{Kullback--Leibler (KL) divergence between predicted and actual distributions of physiological information for each set of modality information.}
\label{fig:KL_to_bio}
\end{figure}

\begin{figure}[t]
    \centering
    \includegraphics[width=0.8\hsize]{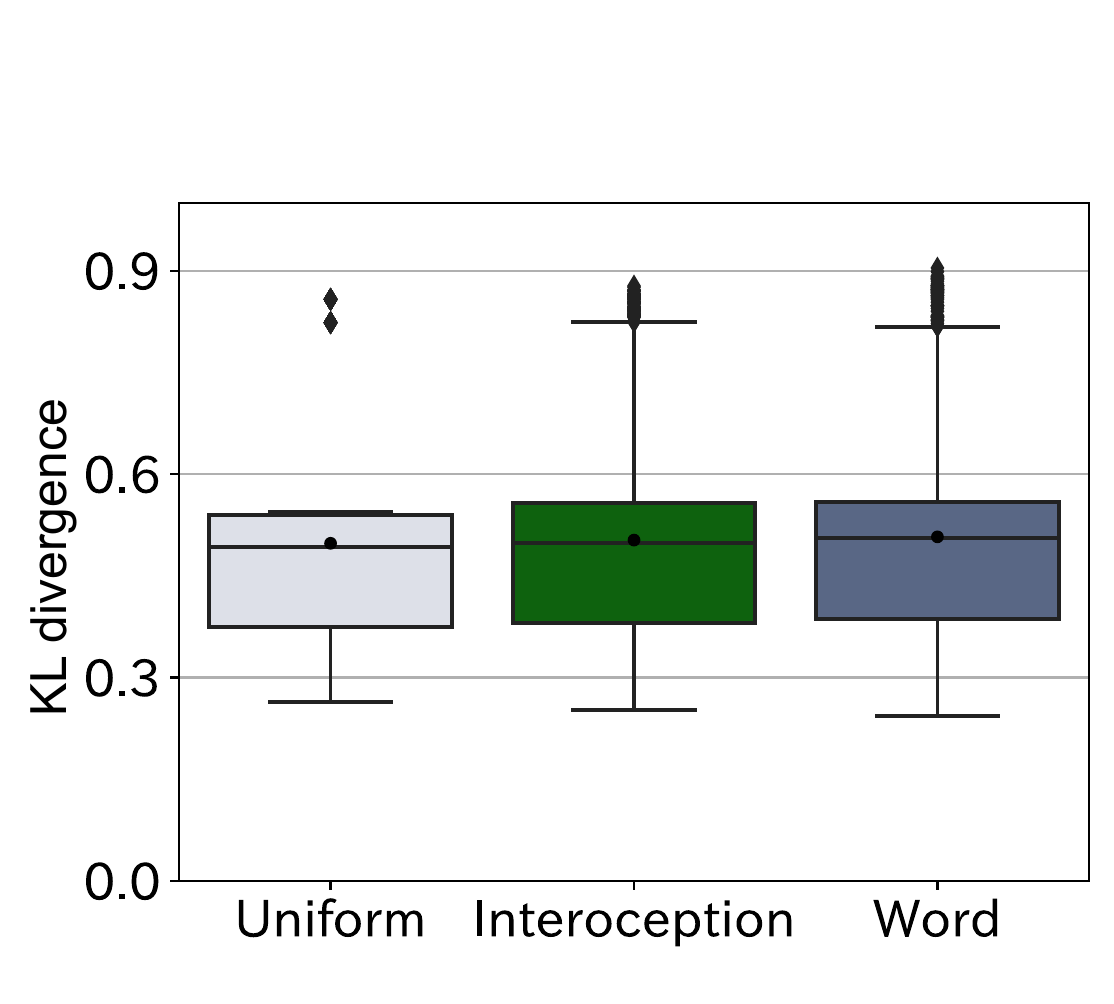}
    \caption{KL divergence between predicted and actual distributions of vision information for each set of modality information.}
    \vspace{-2mm}
    \label{fig:KL_to_image}
\end{figure}


\begin{figure}[t]
\centering
\subfloat[All Words]{\includegraphics[width=2.5in]{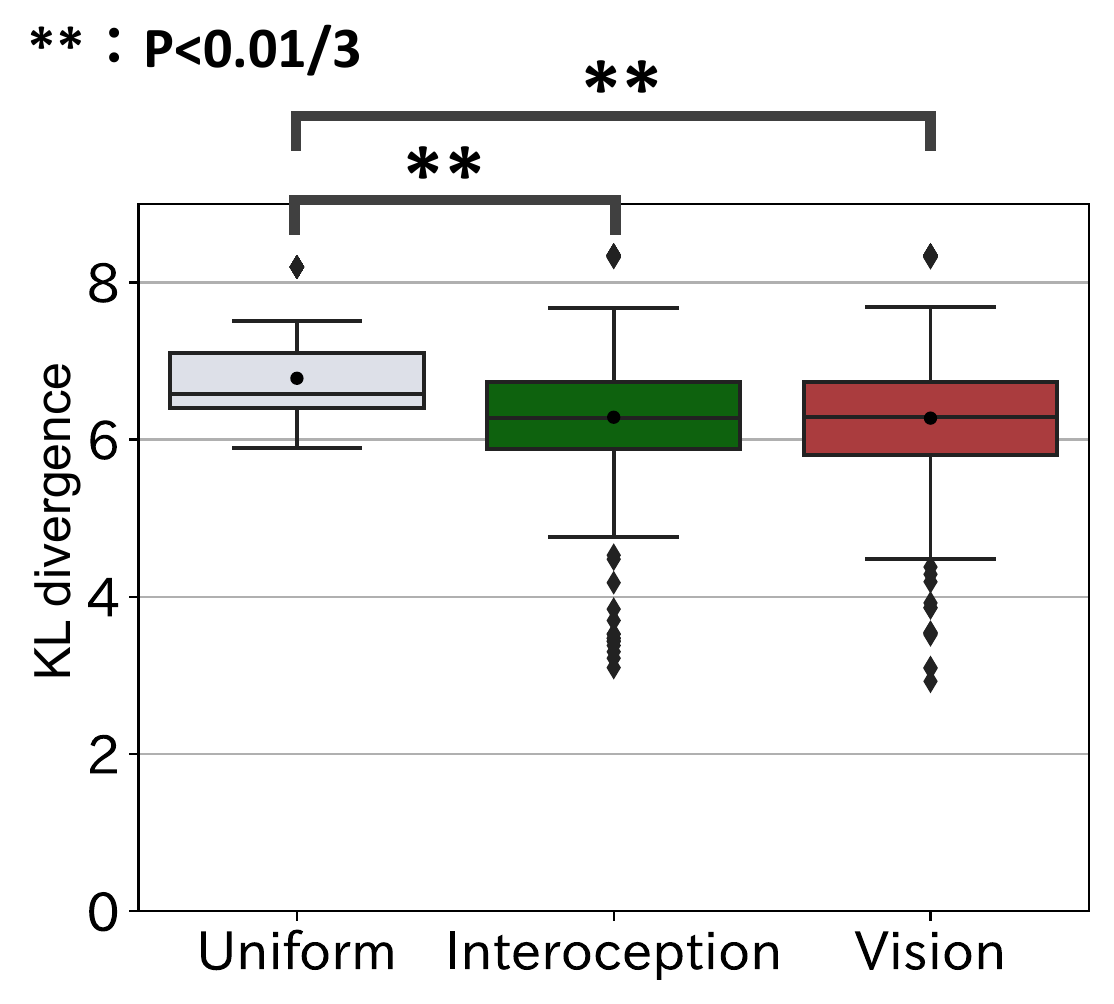}%
\label{fig:KL_to_word_all}}
\hfil
\subfloat[Emotional Words]{\includegraphics[width=2.5in]{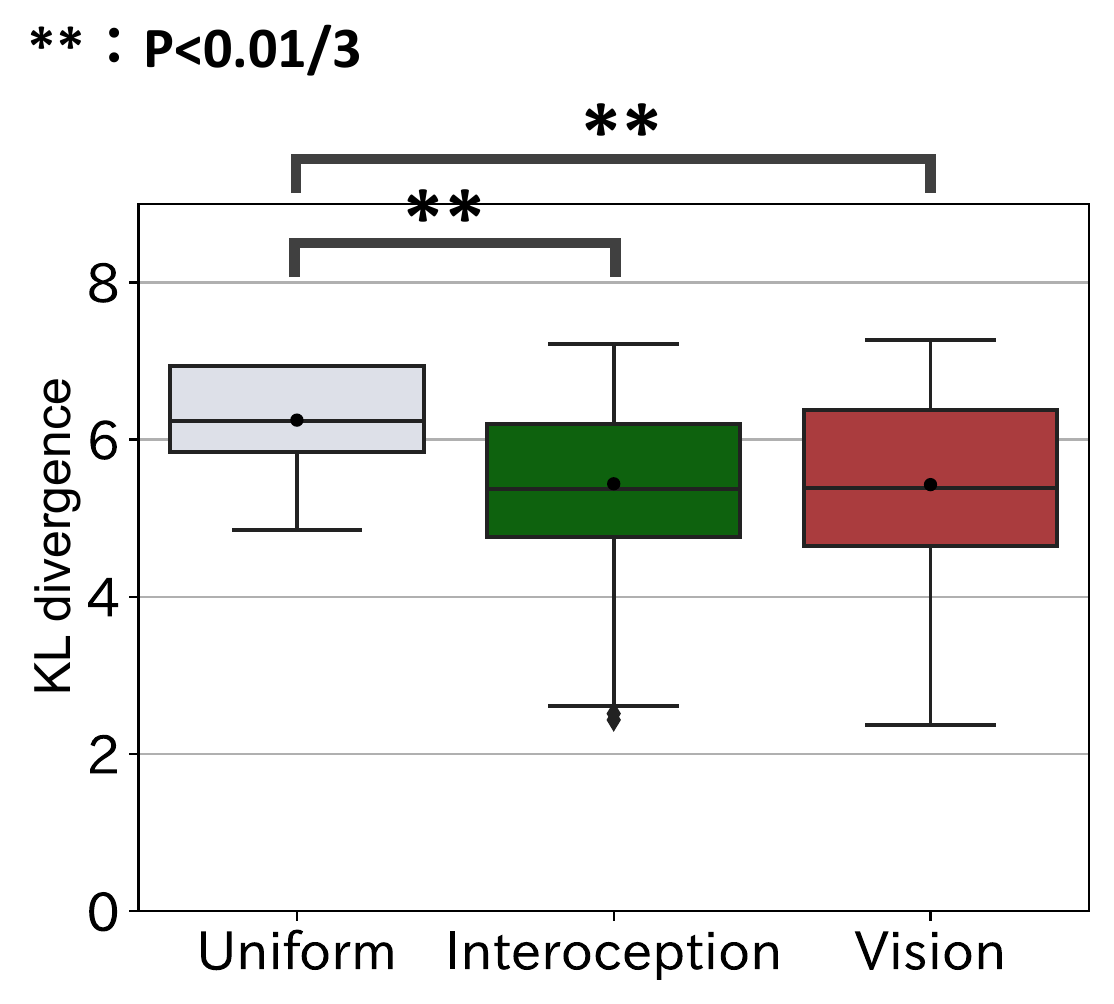}%
\label{fig:KL_to_word_emo}}
\caption{KL divergence between predicted and actual distributions of word information for each set of modality information.}
\label{fig:KL_to_word}
\end{figure}

\begin{table*}[!t]
\caption{Mean Kullback--Leibler (KL) divergence between predicted and actual distributions for each set of modality information.}
\label{tab:KL}
\centering
\begin{tabular}{c|cccccccc}
 \hline
 & \multicolumn{2}{c}{Uniform} & \multicolumn{2}{c}{Interoception} & \multicolumn{2}{c}{Vision} & \multicolumn{2}{c}{Word (All)}\\
 & Mean & SD & Mean & SD & Mean & SD & Mean & SD\\
\hline
\hline
Interoception (EDA) & 3.3 & 0.74 & - & - & 1.9 & 0.69 & 2.0 & 0.60 \\
Interoception (RRI) & 1.9 & 0.69 & - & - & 1.2 & 0.38 & 1.3 & 0.34 \\
Vision & 0.50 & 0.16 & 0.50 & 0.17 & - & - & 0.51 & 0.17 \\
Word (All) & 6.8 & 0.49 & 6.3 & 0.88 & 6.3 & 0.90 & - & - \\
Word (Emotional) & 6.2 & 0.56 & 5.4 & 1.1 & 5.4 & 1.1 & - & - \\
\hline
\end{tabular}
\end{table*}






Next, to determine the prediction accuracy of unobserved information, we used the model trained with learning data, and show the results after comparing the KL divergence for the case of the predicted distribution when unobserved modality information was predicted based on one set of modality information in the test data with the KL divergence for the case of a uniform distribution in Figs. \ref{fig:KL_to_bio}, \ref{fig:KL_to_image}, and \ref{fig:KL_to_word}. The mean and standard deviation values of the results are listed in Table \ref{tab:KL}. For the EDA prediction (Fig. \ref{fig:KL_to_bio}(a)), the KL divergence for the case of the predicted distribution from the vision and word information was significantly smaller than that for the case of a uniform distribution (\(P=6.7 \times 10^{-7}<0.01/3\) and \(P=2.8 \times 10^{-7}<0.01/3\), respectively). Similarly, for the RRI prediction (Fig. \ref{fig:KL_to_bio}(b)), the KL divergence for the case of the predicted distribution from the vision and word information was significantly smaller that for the case of a uniform distribution (\(P=5.9 \times 10^{-5}<0.01/3\) and \(P=2.8 \times 10^{-4}<0.01/3\), respectively). This showed that unobserved physiological information can be predicted using the learned categories. No significant difference was observed between vision and word information in either EDA or RRI (\(P=0.38<0.05/3\) and \(P=0.27<0.05/3\), respectively).

In the vision information prediction (Fig. \ref{fig:KL_to_image}), no significant difference was observed between the predicted distribution from physiology and word information and the uniform distribution (\(P=0.26>0.05/3\) and \(P=0.041>0.05/3\), respectively). Similarly, no significant difference was observed between the physiology and word information (\(P=0.062<0.05/3\)). Therefore, predicting vision information from unobserved data using physiology and word information using this model is likely to be difficult.

For the word information prediction (Fig. \ref{fig:KL_to_word}(a)), the KL divergence for the case of the predicted distribution from physiology and vision information was significantly smaller than that for the case of a uniform distribution (\(P=5.3 \times 10^{-7}<0.01/3\) and \(P=6.7 \times 10^{-7}<0.01/3\), respectively). This shows that unobserved word information can be predicted using the learned model. Moreover, even when reconstructing using only the emotional word results for the prediction results (Fig. \ref{fig:KL_to_word}(b)), the KL divergence was significantly smaller for each when compared to the case of the uniform distribution (\(P=3.0 \times 10^{-6}<0.01/3\) and \(P=4.2 \times 10^{-6}<0.01/3\), respectively). No significant difference was observed between physiology and vision information (\(P=0.21<0.05/3\) and \(P=0.67<0.05/3\), respectively).

\begin{figure}[t]
    \centering
    \includegraphics[width=1\hsize]{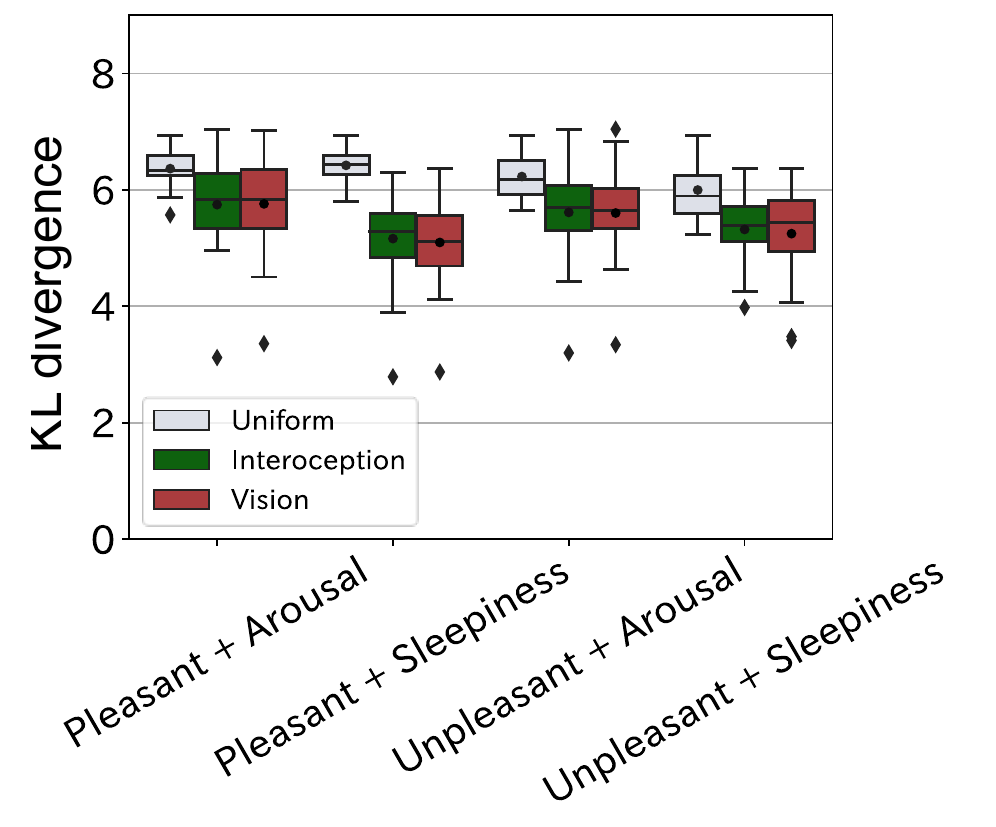}
    \caption{KL divergence between predicted and actual distributions of word information for each set of modality information (only emotional words / category).}
    \label{fig:KL_to_word_emo_by_category}
\end{figure}

Moreover, Fig. \ref{fig:KL_to_word_emo_by_category} shows the prediction results for word information by category, and Table \ref{tab:KL_to_word_emo_by_category} lists the mean and standard deviation values for each category. Fig. \ref{fig:KL_to_word_emo_by_category} shows that the difference from the uniform distribution is greatest in the category of ``pleasant + sleepiness'' for both physiology and vision information. This suggests that the prediction accuracy for images belonging to the ``pleasant + sleepiness'' category was high. Note that statistical analysis was not possible owing to the small sample size for each category.


\begin{table}[!t]
\caption{Mean KL divergence between predicted and actual distributions of word information for each set of modality information (emotional words only / category).}
\label{tab:KL_to_word_emo_by_category}
\centering
\begin{tabular}{c|cccccc}
 \hline
 & \multicolumn{2}{c}{Uniform} & \multicolumn{2}{c}{Interoception} & \multicolumn{2}{c}{Vision}\\
 & Mean & SD & Mean & SD & Mean & SD\\
\hline
\hline
Pleasant + Arousal & 6.4 & 0.31 & 5.7 & 0.74 & 5.8 & 0.76 \\
Pleasant + Sleepiness & 6.4 & 0.29 & \textbf{5.2} & 0.73 & \textbf{5.1} & 0.72\\
Unpleasant + Arousal & 6.2 & 0.38 & 5.6 & 0.75 & 5.6 & 0.75\\
Unpleasant + Sleepiness & 6.0 & 0.47 & 5.3 & 0.60 & 5.2 & 0.78 \\
\hline
\end{tabular}
\end{table}

\section{Discussion}
\label{section:考察}

\begin{figure}[!t]
\centering
\subfloat[EDA\\(Pleasant + Arousal)]{\includegraphics[width=0.5\hsize]{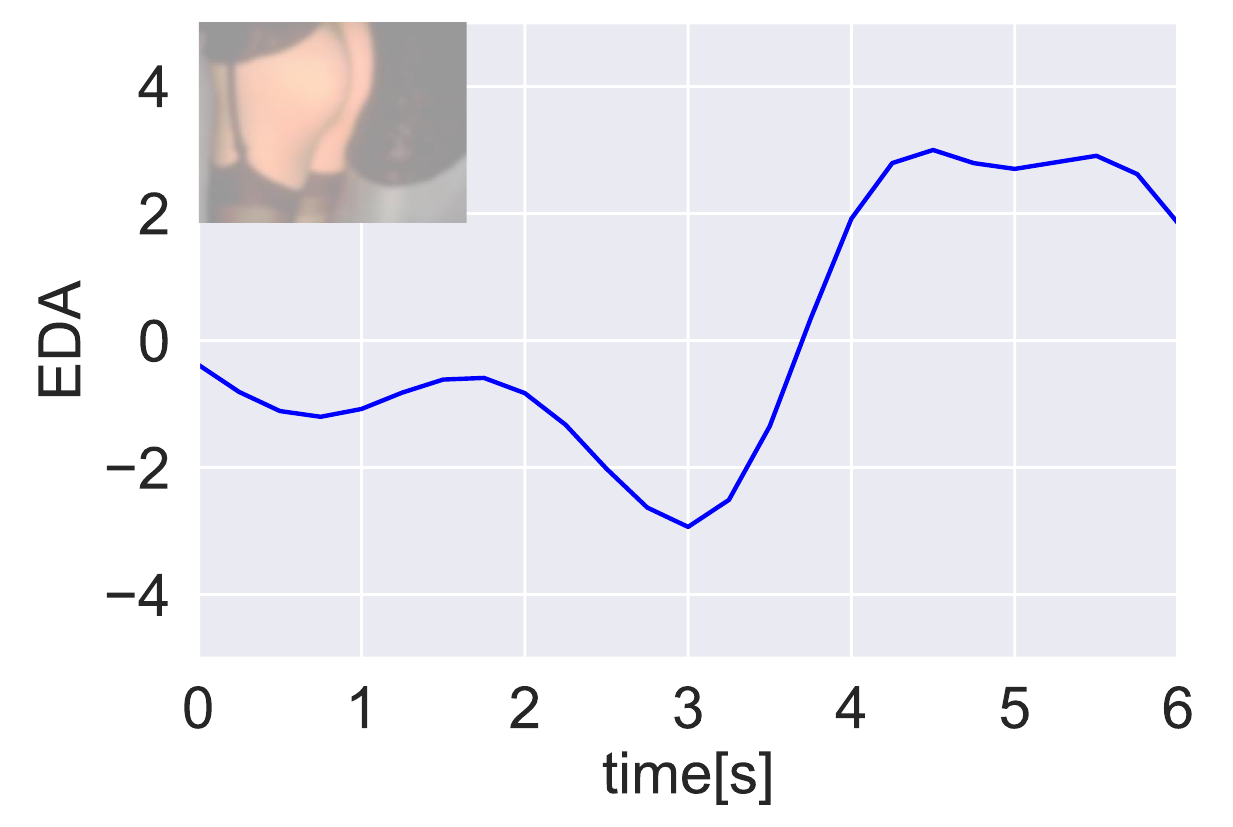}}
\hfil
\subfloat[RRI\\(Pleasant + Arousal)]{\includegraphics[width=0.5\hsize]{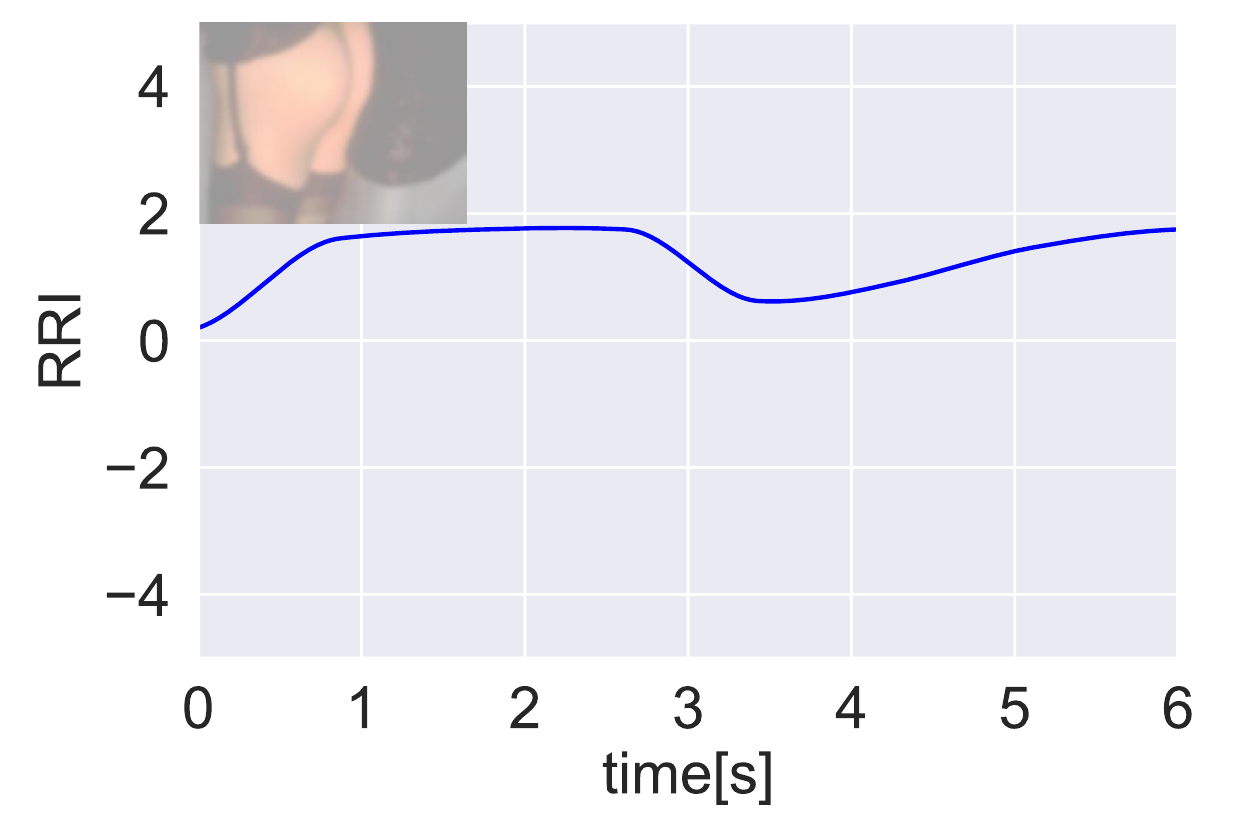}}
\hfil
\subfloat[EDA\\(Pleasant + Sleepiness)]{\includegraphics[width=0.5\hsize]{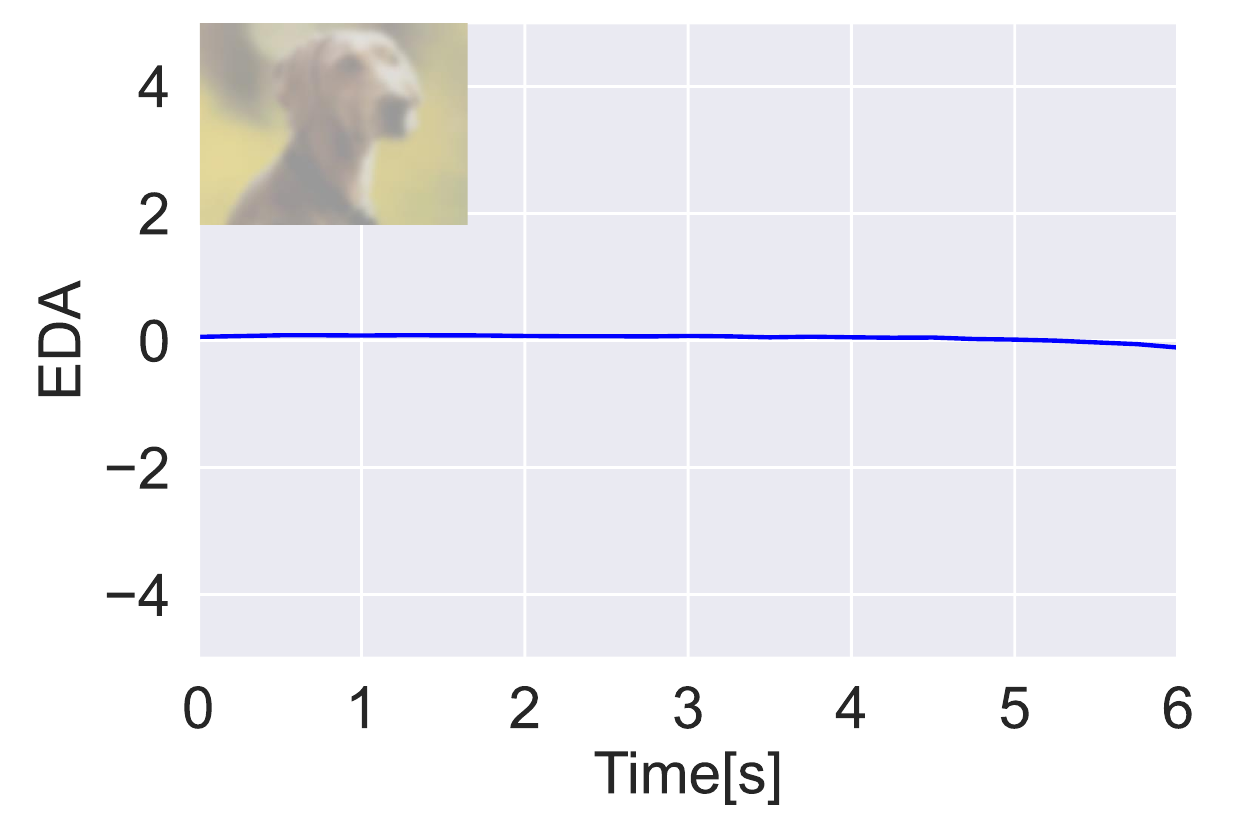}}
\hfil
\subfloat[RRI\\(Pleasant + Sleepiness)]{\includegraphics[width=0.5\hsize]{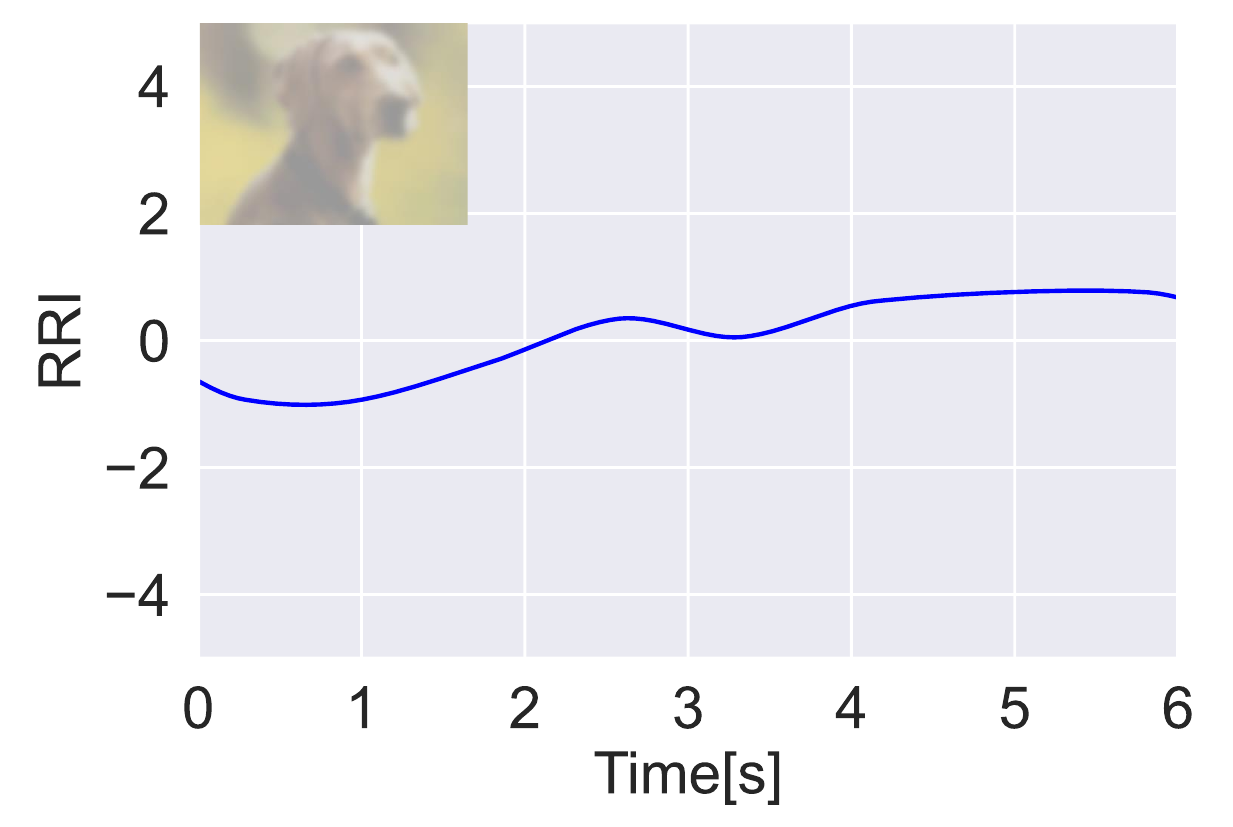}}
\hfil
\subfloat[EDA\\(Unpleasant + Arousal)]{\includegraphics[width=0.5\hsize]{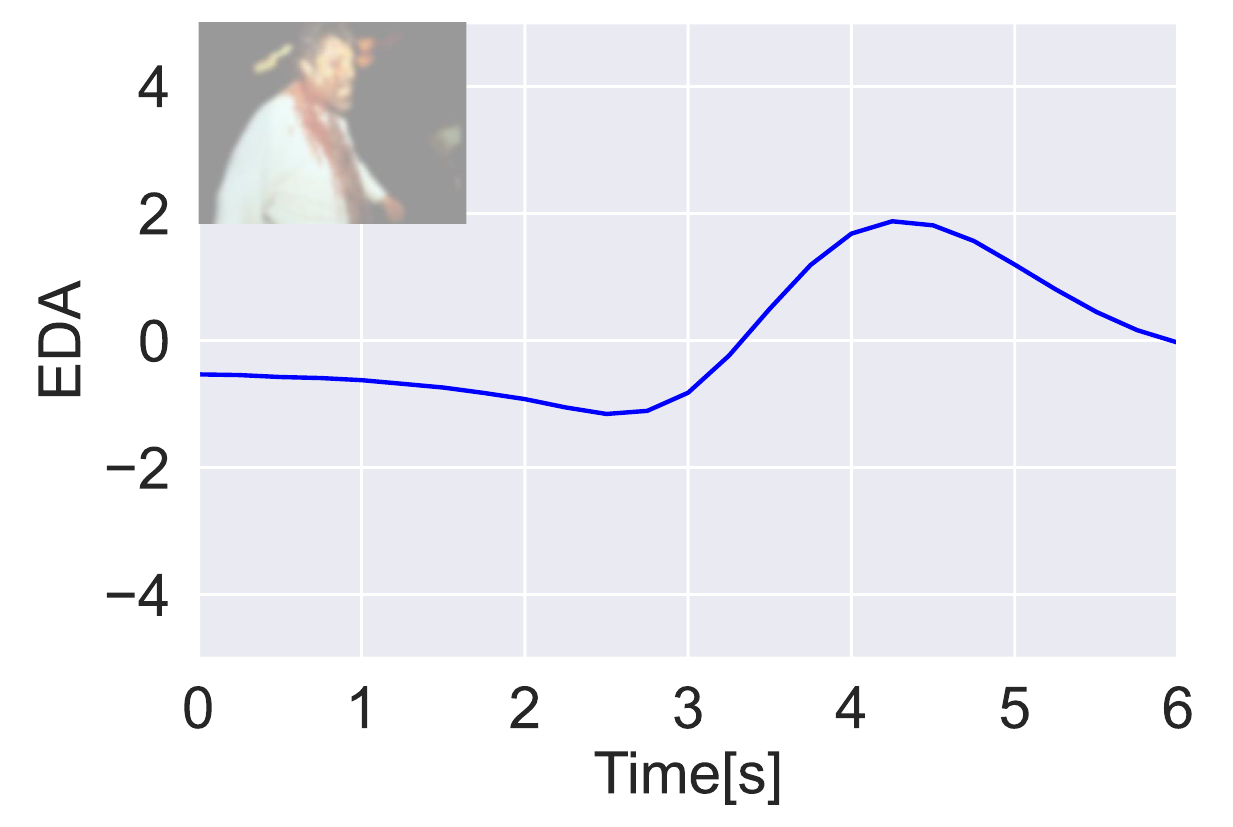}}
\hfil
\subfloat[RRI\\(Unpleasant + Arousal)]{\includegraphics[width=0.5\hsize]{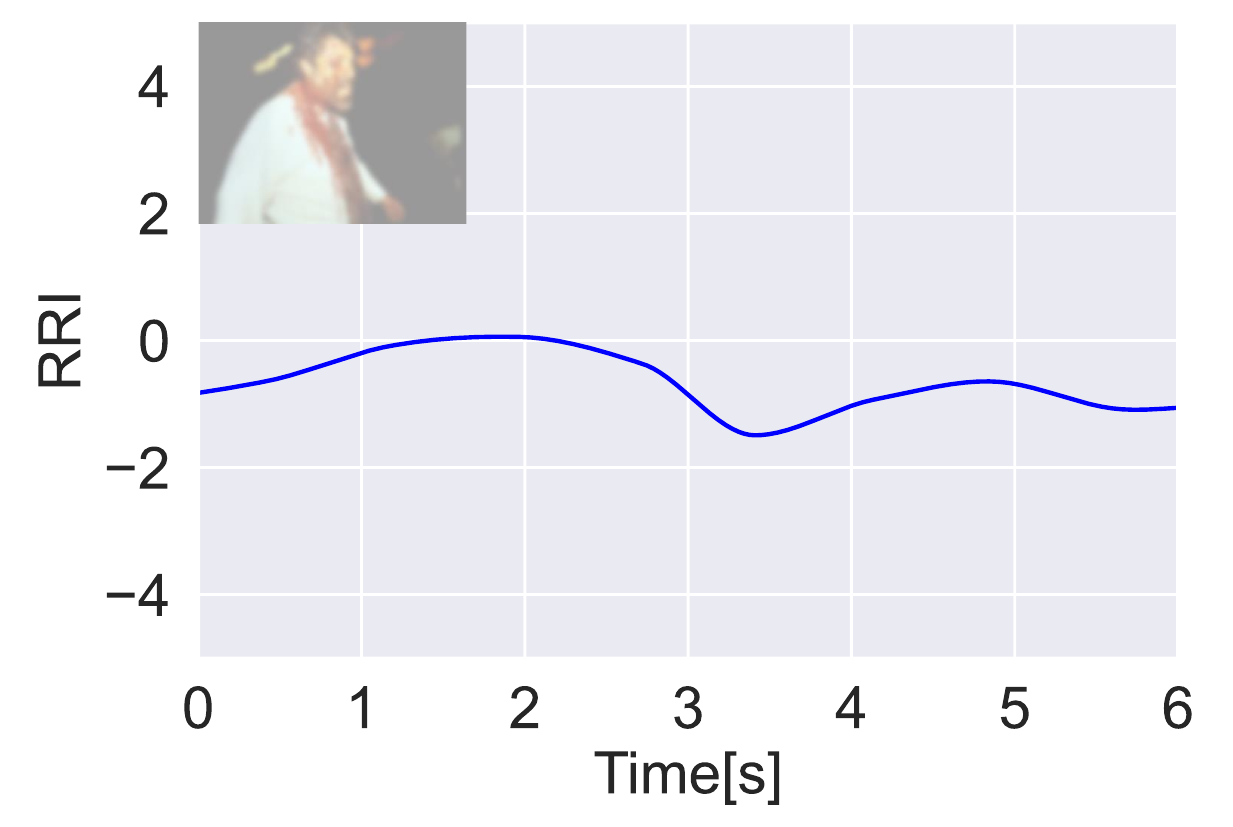}}
\hfil
\subfloat[EDA\\(Unpleasant + Sleepiness)]{\includegraphics[width=0.5\hsize]{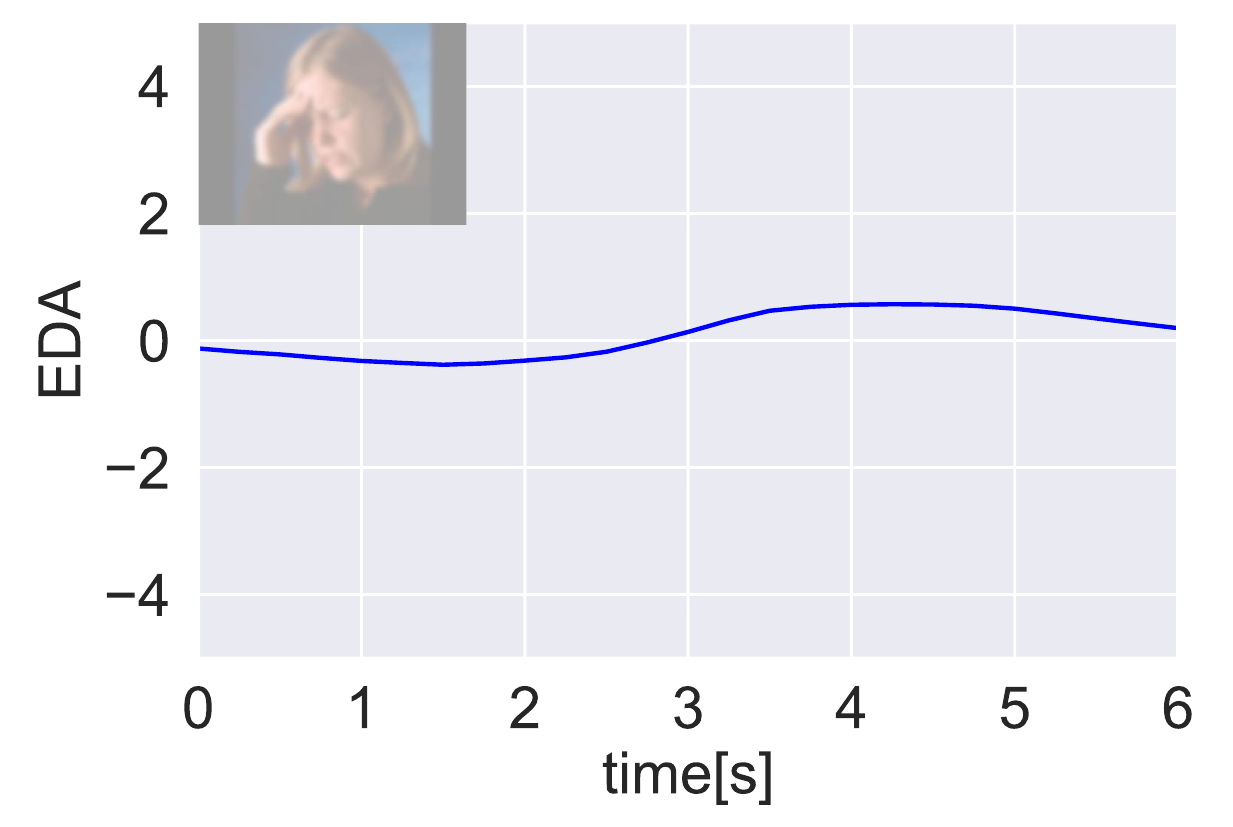}}
\hfil
\subfloat[RRI\\(Unpleasant + Sleepiness)]{\includegraphics[width=0.5\hsize]{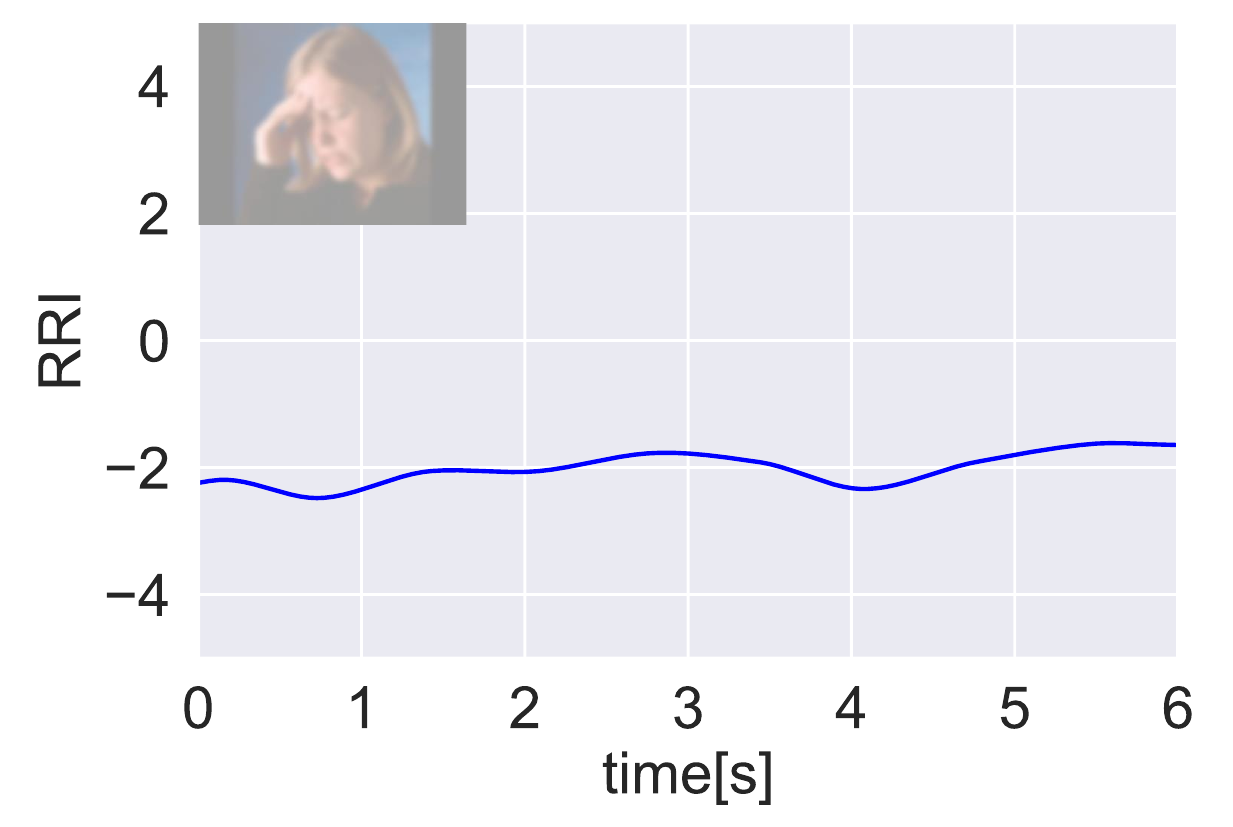}}
\caption{Example of physiological reaction data. IAPS images are blurred due to copyright reasons.}
\label{fig:biosignal}
\end{figure}


In terms of similarity with subjective emotional reports, no significant difference was observed between MLDA and mMLDA.
However, as shown in Fig. \ref{fig:concept}, for certain subjects, the valence and arousal axes clearly appeared in the concept space. When actually investigating the physiological reaction data of the subjects shown in Fig. \ref{fig:concept}, the EDA of the images belonging to the ``unpleasant + arousal'' category at the bottom-right side of Fig. \ref{fig:concept}(a) was elevated as compared to the others (Figs. \ref{fig:biosignal}(a, c, e, g)). EDA is known to have a correlation with arousal level \cite{leiner2012eda}, and this may have resulted in the response in the ``unpleasant + arousal'' category. Moreover, Fig. \ref{fig:concept}(b) shows that the results are clustered to a certain extent in each category. In the case of these subjects, RRI gradually decreased for ``unpleasant + arousal'' and remained flat or gradually increased for the other categories (Figs. \ref{fig:biosignal}(b, d, f, h)). This suggests that the categories were formed based on the magnitude and amplitude of this fluctuation as well as heart rate. The MLDA categories that were affected by these two physiological reactions clustered together by category, as shown in Fig. \ref{fig:concept}(c), and the Rand index was also significantly increased. Furthermore, the addition of vision and word information may have resulted in two-dimensional orthogonal axes appearing as shown in Fig. \ref{fig:concept}(d) as a result of understanding the relationships between categories based on the external environment and past experiences of individuals. The number of subjects in which orthogonal axes appeared in such a concept space was greater in mMLDA than in MLDA, suggesting that both interoceptive and exteroceptive information exhibited an influence on concept formation.

Regarding the prediction of physiology and word information, Figs. \ref{fig:KL_to_bio} and \ref{fig:KL_to_word} show that unobserved physiology and word information can be predicted from other modality information. This was not possible with the MLDA in the present study and is a result that demonstrates the superiority of mMLDA.  

\begin{figure*}[t]
    \centering
    \includegraphics[width=0.8\hsize]{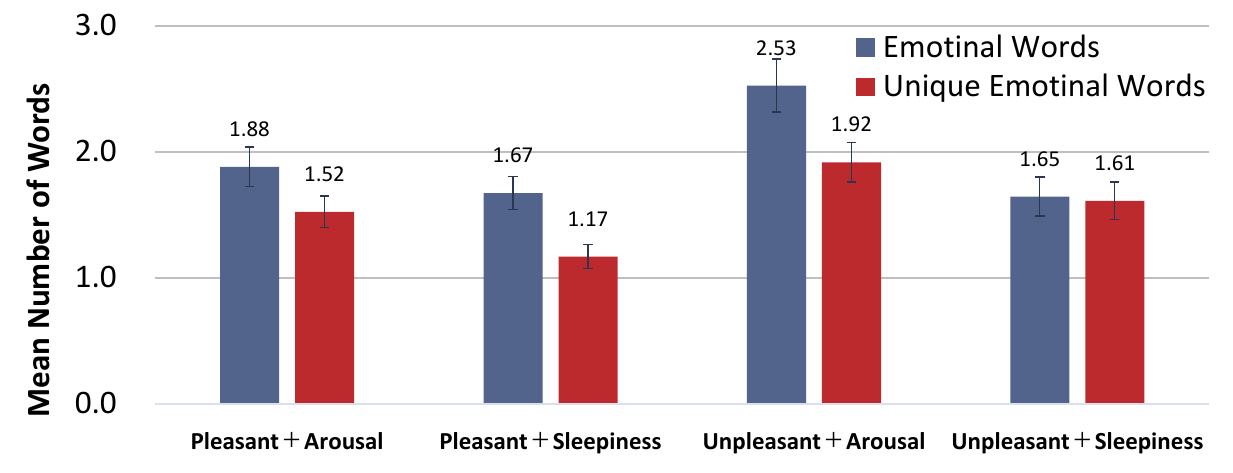}
    \caption{Number of emotional words uttered per image and number of unique emotional words uttered (error bars: standard error).}
    \label{fig:word_number}
\end{figure*}

Furthermore, the results of the KL divergence based on the category in the word information prediction (Fig. \ref{fig:KL_to_word_emo_by_category}) showed that the KL divergence of the ``pleasant + sleepiness'' category was the smallest. Fig. \ref{fig:word_number} shows the numbers of emotional words uttered per image and unique emotional words uttered by category. The figure exhibits a relatively large difference between the number of emotional words uttered per image and number of unique emotional words uttered per image in the ``pleasant + sleepiness'' category. This signifies that each subject repeatedly input the same emotional words for images belonging to the ``pleasant + sleepiness'' category. This may have enabled the learning of the co-occurrence of words and increased the accuracy of word information in unobserved data. However, the accuracy of the ``unpleasant + arousal'' category, which similarly exhibited a large difference between the numbers of emotional words uttered per image and unique emotional words uttered, was not as high as the other categories. As shown in Figs. \ref{fig:biosignal}(e) and (f), the ``unpleasant + arousal'' category exhibited the largest physiological reaction, which may have had an influence; however, in the present study, the amount of data was small, and must be increased to conduct a more detailed analysis.

In this study, 45 images were used as the learning data; however, increasing the sample size may enable more accurate predictions. Moreover, although physiological information could be predicted from vision information, vision information could not be predicted from physiological information. This is speculated to be due to the fact the vision has numerous feature dimensions (1536 dimensions), making it difficult to predict vision information. 
Moreover, the fact that physiological information could be predicted indicates that, although physiological reactions that occur in response to a specific stimulus differ in magnitude and reaction type for each individual, the reaction is relatively consistent within an individual, and the physiological reactions evoked by different visual information can be similar in terms of features.

As these results, the proposed model can be used in robots to acquire emotion concepts as well as objects and places, and is expected to be applied to cases such as understanding the emotions of others in communication situations.

Finally, a limitation of this study is that the individuality of emotion concepts cannot be evaluated. The evaluation in this study involved setting four categories using valence and arousal to compare the subjective evaluation of the emotions of subjects. However, the emotion concepts of an individual are inherently not fixed in nature, but rather are shaped by the experiences of each person. Therefore, not everyone experiences the four categories defined in valence and arousal. Nevertheless, as emotions are shared in society, there are similarities with the evaluation categories set in this study. In the future, we may be able to discuss the different categories each individual possesses using mMLDA to set the number of higher-likelihood emotion concept topics and analyzing each topic. However, currently, no method exits to prove whether this truly matches the emotion concept of each individual; therefore, continued discussion will be needed.

\section{Conclusion}
In this study, we investigated an emotion concept formation model based on an integrated categorization of interoceptive, exteroceptive, and word information in image stimuli. We used an mMLDA model, which was trained using human data when image stimuli were presented. The results showed that the model matched over \(70\%\) of the categories that were subjectively reported by the subjects. This result exceeded the chance level; therefore, we concluded that emotion concept formation can be explained by the proposed model. Moreover, predictions of the unobserved information using a model trained using the learning data exceeded the chance level; therefore, this model could predict unobserved information, which is an important function as a concept. This model can be used to acquire emotion concepts as well as objects and places, and is expected to be applied for the understanding of the emotions of others in communication situations. In the future, we plan to study models related to exteroception other than visual perception and evaluate the individuality of emotion concepts.

\section*{Acknowledgments}
We would like to thank Professor Takayuki Nagai for useful discussions.
This work supported in part by JSPS KAKENHI JP21H04420, JP22H05082 and JST, ACT-X JPMJAX21AL, Japan.
This work involved human subjects or animals in its research. Approval of all ethical and experimental procedures and protocols was
granted by the Research Ethics Committee, Nara Institute of Science and Technology.

\footnotesize
\bibliographystyle{plain}
\bibliography{IEEE_tsurumaki}

\begin{IEEEbiography}[{\includegraphics[width=1in,height=1.25in,clip,keepaspectratio]{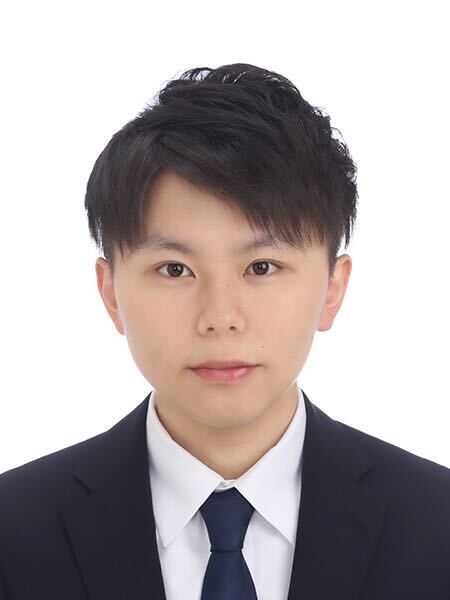}}]{Kazuki Tsurumaki}
received his BE from the Department of Systems Science, Faculty of Engineering Science, Osaka University in 2022. Sinse 2022, he has been an master's student in Systems Innovation, Graduate School of Engineering Science, Osaka University. His research interests emotion model.
\end{IEEEbiography}

\begin{IEEEbiography}[{\includegraphics[width=1in,height=1.25in,clip,keepaspectratio]{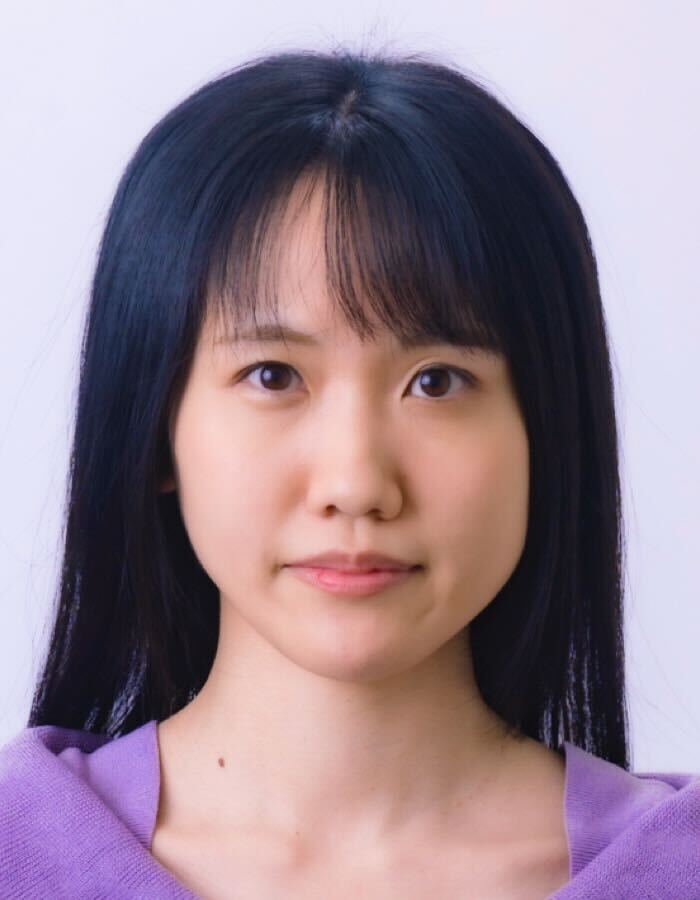}}]{Chie Hieida}
received her BE, ME, and phD degrees from the University of Electro-Communications in 2013, 2015, and 2019, respectively. From 2016 to 2019, she was a research fellow of the Japan Society for the Promotion of Science. From 2016 to 2019, she was a specially appointed researcher at the Symbiotic Intelligent Systems Research Center, Institute for Open and Transdisciplinary Research Initiatives, Osaka University. Since 2020, She has been with the Division of Information Science, Graduate School of Science and Technology, Nara Institute of Science and Technology as a assistant professor. She has received IEEE Robotics and Automation Society Japan Chapter Young Award. Her research is focused on modeling of emotions for robots.
\end{IEEEbiography}

\begin{IEEEbiography}[{\includegraphics[width=1in,height=1.25in,clip,keepaspectratio]{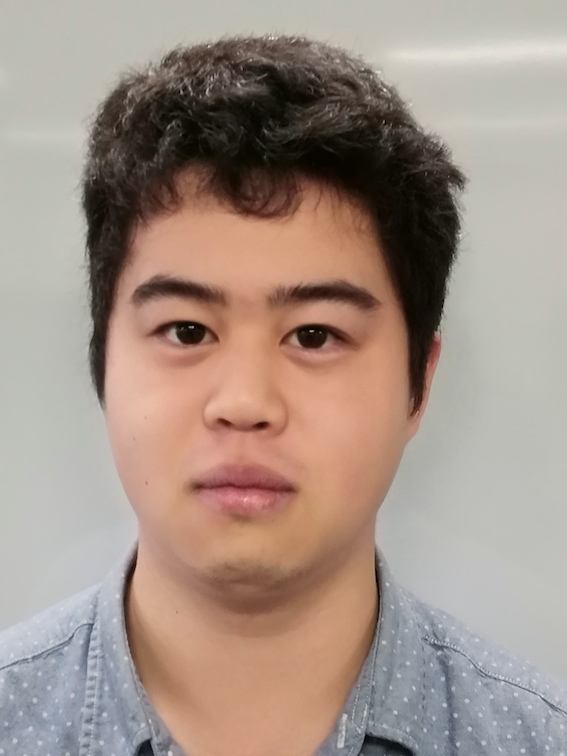}}]{Kazuki Miyazawa}
received Ph.D. degrees from Osaka University, Osaka, Japan in 2022.
He was a JSPS Research Fellowship for Young Scientists from 2019 to 2022.
Since 2022, he has been an Assistant Professor with the Graduate School of Engineering Science, Osaka University.
His current research interests include multimodal data integration, robot learning, concept formation, natural language processing, and reinforcement learning.
\end{IEEEbiography}


\end{document}